\documentclass[sigconf]{acmart}

\usepackage{tablefootnote}
\usepackage[subrefformat=parens]{subcaption}
\usepackage{acmart-taps}
\usepackage{amsmath}
\usepackage{bbm}
\usepackage{enumitem}

\AtBeginDocument{%
  \providecommand\BibTeX{{%
    \normalfont B\kern-0.5em{\scshape i\kern-0.25em b}\kern-0.8em\TeX}}}

\copyrightyear{2025}
\acmYear{2025}
\setcopyright{cc}
\setcctype{by}
\acmConference[CHI '25]{CHI Conference on Human Factors in Computing
Systems}{April 26-May 1, 2025}{Yokohama, Japan}
\acmBooktitle{CHI Conference on Human Factors in Computing Systems (CHI
'25), April 26-May 1, 2025, Yokohama,
Japan}\acmDOI{10.1145/3706598.3713745}
\acmISBN{979-8-4007-1394-1/25/04}

\newcommand{\squishlist}{\begin{itemize}[itemsep=1pt,parsep=2pt,topsep=3pt,partopsep=0pt,leftmargin=0em, itemindent=1em,labelwidth=1em,labelsep=0.5em]}
\newcommand{\squishend}{\end{itemize}}
\newcommand{\xref}[1]{\S\ref{#1}}

\begin{document}

\title{Spatial Speech Translation: Translating Across Space With Binaural Hearables}
 \author{Tuochao Chen}
 \affiliation{Paul G. Allen School,  University\\ of Washington, Seattle, WA  
 \country{USA}
 }
 \email{tuochao@cs.washington.edu }

 \author{Qirui Wang}
 \affiliation{Paul G. Allen School, University\\ of Washington, Seattle, WA
  \country{USA}
 }
 \email{qw43@cs.washington.edu}

  \author{Runlin He}
 \affiliation{Paul G. Allen School, University\\ of Washington, Seattle, WA
  \country{USA}
 }
 \email{rh74@cs.washington.edu}

 \author{Shyamnath Gollakota}
 \affiliation{Paul G. Allen School, University\\ of Washington, Seattle, WA
   \country{USA}
 }
\email{gshyam@cs.washington.edu}

\renewcommand{\shortauthors}{Chen, Wang, He and Gollakota}

\begin{abstract}
Imagine being in a crowded space where people speak a different language and having hearables that transform the  auditory space into your native language, while preserving the spatial cues for all speakers.  We introduce  {\it spatial speech translation}, a novel concept for hearables that translate  speakers in the  wearer's environment,  while maintaining the direction and unique voice characteristics of each speaker in the binaural output. To achieve this, we tackle several technical challenges  spanning blind source separation, localization, real-time expressive translation, and binaural rendering to preserve the speaker directions in the translated audio, while achieving real-time  inference on the Apple M2 silicon. Our proof-of-concept evaluation with a prototype binaural headset shows that,  unlike existing models, which fail in the presence of interference, we achieve a BLEU score of upto 22.01   when translating between languages, despite strong interference from other speakers in the environment.  User studies further confirm the system’s effectiveness in spatially rendering the translated speech in previously unseen real-world reverberant environments. Taking a step back, this work marks the first step towards integrating spatial perception into  speech translation. \noindent{Code, dataset available at {\textcolor{blue}{{{\url{https://github.com/chentuochao/Spatial-Speech-Translation}}}}}}

\end{abstract}

\ccsdesc[500]{Computing methodologies~Machine translation}
\ccsdesc[500]{Human-centered computing~Sound-based input/output}
\ccsdesc[500]{Human-centered computing~Ubiquitous and mobile computing}
\ccsdesc[500]{Computing methodologies~Machine learning}

\keywords{Speech translation, spatial computing, augmented audio}

\begin{teaserfigure}
\centering
  \includegraphics[width=1\textwidth]{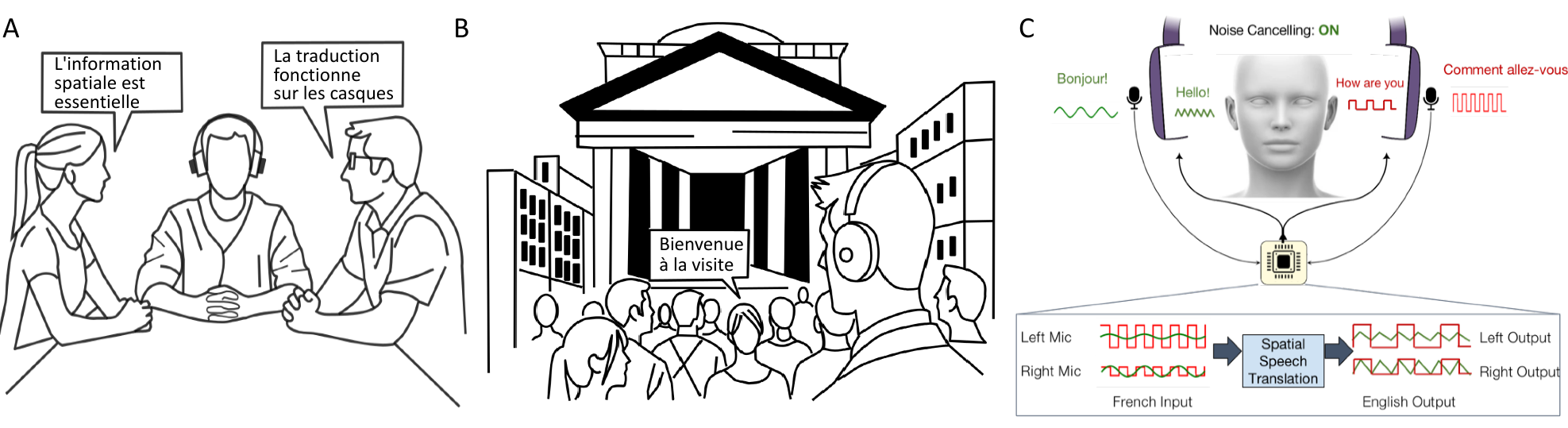}
  \vskip -0.1in
  \caption{"Spatial speech translation" is an intelligent hearable system that translates speakers in the wearer's auditory space, preserving the direction and unique voice characteristics of each speaker in the binaural output. (A) Two speakers have a conversation, and the wearable translates both in real-time, while maintaining their spatial and acoustic features. (B) In a crowded environment, the hearable uses binaural cues for directional translation, translating only the speaker from a specific direction (e.g., where the wearer is looking) and ignoring other speakers in the environment. (C) The noise-canceling headset captures binaural input, processes the signals, and plays back the translated binaural speech in real time.}
  \label{fig:teaser}
\end{teaserfigure}

\maketitle


\section{Introduction}

Since ancient times, language translation has served as a crucial tool for the exchange of knowledge and cultural practices~\cite{classic}. More recently, science fiction novels, television shows, and films have popularized the idea of wearable devices capable of seamless speech-to-speech translation, exemplified by the fictional Babel fish from the Hitchhiker's Guide to the Galaxy and the  universal speech translator from Star Trek~\cite{startrek-npr}. In this paper, we explore a novel concept for hearables — {\it spatial speech translation} — that allows wearers  to 
 translate speakers in the
auditory space seamlessly and in real-time, while preserving the spatial cues as well as  the unique voice characteristics of each individual speaker in the translated binaural output.

To appreciate the potential applications, consider arriving in a small town in France, without the ability to speak french. As one navigates the city, people talk on phones, chat in groups, and laugh together. The language barrier can feel like a virtual wall, leading to a sense of isolation. At the station, an announcement prompts reactions among the crowd—some move to different platforms, while others exchange information. The words are a blur—you can hear the chatter around you and pick up on the rhythm of the voices, but the meaning remains unclear. Now, imagine a hearable device that translates multiple speakers within the wearer's auditory space (Fig.~\ref{fig:teaser}A) or even a single speaker from a specific direction (Fig.~\ref{fig:teaser}B). Such a device would enable the wearer to understand surrounding conversations. If the device also preserves the direction, prosody, and vocal characteristics of each speaker in real-time, the wearer can seamlessly identify who is saying what in  multi-speaker settings. Such a technology could transform the wearer's  auditory space into their native language while maintaining the unique voices and spatial orientation of the translated speakers.


The above technology, we call spatial speech translation, is a new capability for hearable devices. While some mobile devices now offer translation~\cite{pixelbuds,timekettle}, they neither support translating multiple speakers within the wearer's environment nor incorporate spatial awareness in either the audio input or the  translated  output. To our knowledge, our work is the first to bring spatial perception to the problem of speech translation.  Achieving this requires multiple capabilities spanning source separation, localization,  translation and binaural  audio rendering. 

Specifically, our system has  the following key requirements.

\squishlist

\item {\bf Speaker localization and separation.}
We need to  estimate the number of audible speakers in the wearer's auditory space, e.g.,  in Fig.~\ref{fig:teaser}A, without prior knowledge of the number of speakers. We must also compute the direction of each speaker and separate them into individual  speech signals using the binaural signals from the hearables. The separation and localization algorithms must  be robust to variations in head-related transfer functions (HRTFs) across  wearers.

\item {\bf Simultaneous speech translation.} Human speech and translation are attuned to nuances like turn-taking and timing~\cite{Levinson2016TurntakingIH}. Simultaneous human interpreters, for example,  balance speed and accuracy --- delaying too much  disrupts communication, while rushing results in subpar  translation quality as important information may be missed~\cite{seamlessexpressive,streamspeech}. Instead of waiting for a whole speech utterance to finish before translating, simultaneous translation requires real-time processing based on local context to minimize translation delays to a few seconds,  while maintaining high accuracy.

\item{\bf Expressive speech translation.} Preserving vocal style in translated speech involves capturing prosodic elements such as pitch, stress, and rhythm, which convey meaning, emotion, and intent. Maintaining these vocal characteristics helps the wearer better associate the translated speech with the original speaker, especially when multiple speakers are present in the auditory space.

\item {\bf Binaural rendering of translated audio.} The translated speech from  speakers in the wearer's auditory space must be played back through the hearables, while preserving binaural cues and the direction of each speaker. We can also choose to translate and render only  a single speaker  from a specific direction in which the wearer is looking. Maintaining the direction and incorporating spatial awareness is crucial for an accurate and immersive experience.
\squishend

We address these challenges and demonstrate spatial speech translation. We make three key technical contributions.

\squishlist
\item {Existing translation systems fail when interfering speakers are present (see~\xref{sec:rel}). Addressing the challenges of multiple speakers and interference requires neural network-based source separation~\cite{acousticswarm,cone-of-silence}, which has not yet been explored for speech translation.} To this end, we designed a search-based method that utilizes binaural headsets  to perform real-time joint localization and separation, outputting a binaural signal for each separated speaker. We divide the auditory space into small angular regions, and within each region, a neural network searches for potential speakers. The model is trained to  extract a speech source if one is present and to output silence if no speech is detected. We  show that our training methodology enables  generalization  to real-world reverberations and multipath effects, as well as variability in head-related transfer functions (HRTFs) across wearers, without the need to collect any training data, for each language, with our prototype hardware.
\item {Existing simultaneous speech-to-speech translation models either are too large for on-device use~\cite{seamlessexpressive} or lack expressive translation~\cite{streamspeech}.} We design and train a simultaneous, expressive speech translation model capable of  running in real-time on Apple silicon. We also make the observation that the output from our separation model, which  feeds into the translation model,  contains residual distortions from interfering speakers and noise. This can reduce translation performance since the model has not been trained on such distortion.  To enhance robustness, we  fine-tune the translation model using the output of our separation model and demonstrate improved translation performance (see~\xref{sec:translate:design}).
\item  {Binaural rendering, which is important for spatial audio~\cite{hrtfuist}, has not been explored in prior work for real-time speech translation. } This requires creating the perception that the translated speech originates from the  direction of the corresponding speaker. We present a method that transfers spatial cues from the binaural input to the translated output. At a high level, the joint localization and  separation algorithm described above provides binaural separated speech signals, along with the corresponding direction for each speaker.  Using this, we estimate spatial cues for each speaker and apply them to their  translated speech, while accounting for the delay inherent to  translation models.

\squishend

To demonstrate proof of concept, we used an off-the-shelf noise-canceling headset (Sony WH-1000XM4) and commercial wired binaural earphones (Sonic Presence SP15C) with access to both microphones. Our models were trained to translate from {French, German and Spanish} to English and were implemented on Apple M2 silicon, { which is supported by commodity devices including augmented reality (AR) headsets like the Apple Vision Pro.} A runtime inference analysis is presented in \xref{sec:implementation}. Our empirical results for our French-English system are as follows:

\begin{figure*}[t!]
\centering
  \includegraphics[width=0.7\textwidth]{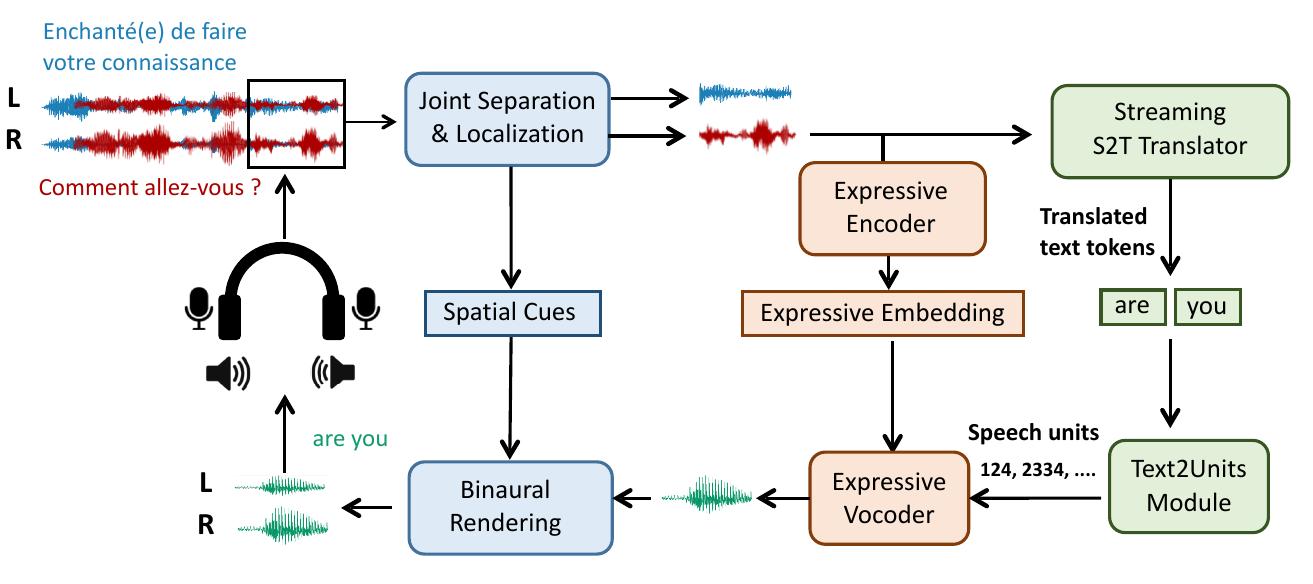}
  \vskip -0.1in
  \caption{{\bf Overview of spatial speech translation.}
The input to our pipeline is a binaural noisy speech mixture in the source language (e.g., French). It consists of three main components: 1)  A lightweight, streaming model that separates and localizes individual speech within the binaural mixture, extracting spatial cues for each voice. 2)  A streaming speech translation model that translates the separated speech chunks into the target language (e.g., English) while an expressive encoder and vocoder preserve the vocal qualities and expressiveness of the original audio. 3) Binaural rendering to reconstruct  binaural playback using the extracted spatial cues. }
  \vskip -0.15in
  \label{fig:diag}
\end{figure*}

\squishlist

\item   Real-world evaluations with  participants wearing our binaural headsets with  concurrent  speakers in {six}   different indoor and  {four outdoor} scenarios showed  generalize to previously unseen participants, and multipath environments. Our design did  not require any   training data collection with our hearable hardware.
\item In a user study with {29}  participants who spent over 350 minutes rating translation outputs from real-world environments, our system achieved a semantic consistency score of {3.35}, computed between the ground truth and translated speech. In comparison, existing models without spatial awareness scored {1.15}. Additionally, our fine-tuning method for robustness to source separation improved the BLEU score from {18.06 to 22.07}. 
\item In the same user study, our system improved opinion scores for speaker similarity between the ground truth  speech and the translated English output  from {1.81 to 3.45}. Additionally, the objective Vocal Style Similarity (VSim) metric improved from {0.013} in the non-expressive model to {0.250} in our expressive model.
\item In a spatial hearing study, participants predicted the  direction for both original  and translated English samples with similar  errors. Our  rendering method reduced interaural time difference ($\Delta$ITD) and interaural level difference ($\Delta$ILD) errors to 72.3~µs and 0.16~dB, respectively, compared to 314.9~µs and 1.83~dB without our method.
\squishend


{{\bf Contributions.} This paper introduces {\it Spatial Speech Translation,} the first real-time binaural hearable system that translates speech from multiple speakers in the wearer’s auditory space while preserving spatial cues and the unique voice characteristics of each speaker. Our work makes several key contributions:}

\squishlist  
\item {{\it On-device simultaneous, expressive speech translation: } We design and train the first simultaneous and expressive speech translation model that runs in real-time on Apple silicon, enabling its use on commodity wearable devices.}

\item  {{\it Integrating speech translation with localization and source separation: } While existing translation systems fail in the presence of interference, we use a search-based method that leverages binaural input to perform real-time joint localization and source separation.}

\item {{\it Robustness to  separation  imperfections:} The output of the separation network is inherently imperfect and may contain residual distortions from interfering speakers and noise, which degrade translation performance. To address this, we propose a fine-tuning technique that trains the translation model on these residual distortions and interference, significantly improving its robustness.}

\item {{\it Binaural rendering of translated speech:} We propose and compare three different methods to transfer spatial cues from the binaural input to the translated output, achieving the first binaural rendering of speech translation on a real-time hearable system.} 

\item {{\it Human-centric evaluation:}  Through user studies conducted in both indoor and outdoor environments, we provide quantitative and qualitative evaluations of our spatial translation system, situating the results in the context of human experience and perception.}
\squishend



\section{Background and related work}\label{sec:related}

To the best of our knowledge,  prior work on intelligent hearable systems has not explored the concept of spatial speech translation. Below we describe related work in mobile translation, speech processing and  deep learning.

\vskip 0.05in\noindent{\bf Mobile translation.} Early websites, such as AltaVista Babelfish, played a notable role in popularizing text-based machine translation~\cite{nprbabelfish,humanfactors-1,humanfactors-2}. Modern translation platforms continue to utilize the "two text box" interface initially developed by SYSTRAN~\cite{humanfactors-2,huamnfactors-3}. In the last decade, numerous smartphone translation apps have emerged, retaining this interface while incorporating speech recognition technology. For instance, the Google Translate app features a conversational mode where users press a button to speak, and their speech is simultaneously translated to text but only read aloud in the target language once they finish speaking. These systems support simultaneous speech-to-text translation but lack the capability for simultaneous speech-to-speech translation. Moreover, prior studies have shown that the need to press buttons and pass devices between users disrupts conversational flow, introduces latency, and interferes with non-verbal communication cues present in face-to-face interactions~\cite{humanfactors-2}. Consequently, researchers have suggested the need for a touch-free translation experience featuring an eyes-free interface~\cite{humanfactors-1,humanfactors-2}.

A few commercial  translation devices have also been introduced~\cite{pixelbuds,timekettle,manu,waverly}. Google's Pixel Buds  enable one user to wear earbuds while the other holds a phone. Similarly, Timekettle WT2 Edge earbuds~\cite{timekettle} offer translation by sharing the earbuds between users, i.e., each user uses a single earbud. These devices have multiple key limitations: 1) they fail to preserve spatial cues in the translated output, which are essential for effective human communication, 2)  they do not support expressive translation and hence prosody cues and the speaker characteristics are lost in translation, and  3) they only support bi-directional translation between two people and lack uni-directional translation capabilities for translating multiple speakers in a space; they can translate only one target speaker at a time, identified through shared use of the phone or earbuds. In contrast, we introduce  spatial speech translation where speakers in the wearer's space seamlessly gets translated, while preserving the spatial and prosody cues of each individual speaker.


\vskip 0.05in\noindent{\bf Speech processing in noisy environments.} The advent of deep learning has transformed  speech  research, with  neural networks demonstrating superior performance to traditional signal processing methods for numerous audio  tasks~\cite{better1,better2,semantichearing,soundbubble,tce}. A common task  is to enhance a speech signal in the presence of noise, i.e., speech enhancement, where the goal is to denoise a speech signal from a mixture of sounds~\cite{clearbuds,denoise1,denoise2}. Directional hearing  has  been proposed to extract a speech signal using directional cues from multiple   microphones~\cite{directionalhearing1}. Target speech hearing systems  have also been proposed to extract a speaker based on their speech characteristics and distance~\cite{tsh-chi24,tse1,tse2,tse3,tse4,soundbubble}. The work closest to ours is prior research on blind source separation~\cite{bse}, where the goal is to separate a mixture of speakers  into individual audio streams. Previous work has used multiple microphones to perform blind source separation and estimate the angular positions of each speaker~\cite{cone-of-silence,acousticswarm,romit1}.  Building on this work, we design the first spatial speech translation system that performs both blind source separation and translation of each individual speaker, while preserving the direction of the speakers in the translated binaural output.


\vskip 0.05in\noindent{\bf Neural networks for translation.} Machine translation is a classic problem in natural language processing where the goal is  automatic translation of written text from one natural language to another~\cite{machinetranslate,Bahdanau2014NeuralMT}. Similar to speech separation, neural networks and in particular the transformer architecture have  demonstrated  significant performance improvements for machine translation~\cite{attentionpaper}. Speech translation is a related task, but its performance has lagged behind that of machine translation~\cite{seamlessexpressive}. Here the objective is to translate speech from one language to either speech or text in another language~\cite{takuya1,endtoend1,endtoend2,endtoend3}. The traditional approach to implementing speech translation has been to cascade an automatic speech recognition (ASR) model that converts speech into text, which is then input into a machine translation model~\cite{cascade1,cascade2,cascade3}. However, such cascaded approaches are limited as non-linguistic information like prosody, expressiveness, pauses, and speaker characteristics are lost in text-based input, making it difficult to restore them in the translated speech output~\cite{seamlessexpressive}. Further, the delay and computational complexity of  a cascaded system can be higher since the machine translation system has to wait for the ASR output~\cite{takuya1}. 

This has led to the recent interest in creating end-to-end neural speech-to-speech translation that have demonstrated superior performance to cascaded systems~\cite{endtoend1,endtoend2,endtoend3,endtoend4,wang2023neuralcodeclanguagemodels,Lee2021DirectST}. So far, two key research directions have been considered for speech translation systems:  1) real-time simultaneous  translation that minimizes latency without compromising translation quality,  and 2)  expressive translation that preserves the vocal styles and prosody of the speaker in the translated speech.  In our paper, we introduce a third research direction which is the spatial aspect of speech translation. 

\vskip 0.05in\noindent{\bf Simultaneous translation.} Recent works have built  large speech translation models that can achieve simultaneous translation~\cite{seamlessexpressive,google-simul,simulspeech,llmtrans,ctc}. These models either use rule-based policies~\cite{rule1,rule2,rule3,rule4,rule5,simulspeech} or learnable policies~\cite{learn1, learn2,learn3} to wait for more input before translating. Rule-based policies rely on heuristics like waiting for a fixed number of tokens to be heard before translating, while learnable policies, are closer to human translators, in that they use local context and learnable parameters to make this decision~\cite{seamlessexpressive}. These works however do not consider the spatial aspect of speech translation and are limited to a single speaker.

\vskip 0.05in\noindent{\bf Expressive speech systems.} Text-to-speech (TTS) synthesis systems can achieve voice style transfer using techniques like flow matching~\cite{voicebox}, diffusion models~\cite{shen2023naturalspeech2latentdiffusion} and speech language models~\cite{wang2023neuralcodeclanguagemodels}. This led to interest in creating expressive speech translation models~\cite{seamlessexpressive}. TTS systems can achieve cross-lingual transfer when
stacked with translation models~\cite{borsos2023audiolmlanguagemodelingapproach,audiopalm,polyvoice}. In late 2023, Meta released a speech-to-speech translation system that achieves both simultaneous  translation and sentence-level voice style preservation~\cite{seamlessexpressive}. Currently,  however, there is no trainable open-source model that supports both these features. While the Meta model~\cite{seamlessexpressive} is available, it is large (around 2 billion  parameters) and does not provide the  training script, which prevents it from being modified to create a spatial model. In contrast,  we  introduced  the concept of spatial speech translation as well as  developed an open-source trainable model that supports real-time, simultaneous and expressive speech translation that may be useful to the research community.

\section{System Design}

Fig.~\ref{fig:diag} shows a block diagram of the different components in our system. At a high level, we have three key components: joint localization and speech separation, real-time expressive translation with finetuning and binaural rendering for translated speech. 

\subsection{Joint localization and speech separation}\label{sec:jls}
This step has two related goals: (1) To translate the wearer's acoustic environment, we need to estimate the number of speakers and extract individual speech signals for each to enable translation, and (2) we need to localize and determine the angle of arrival (AoA) for each speaker to preserve the spatial properties of the input speech in the translated output. Inspired by \cite{cone-of-silence, acousticswarm}, we adopt a search-based joint localization and separation approach for our binaural setup. Compared to blind separation, search-based methods have shown better performance, especially when the number of speakers in the acoustic space is unknown.

\begin{figure*}[t!]
\centering
  \includegraphics[width=0.7\textwidth]{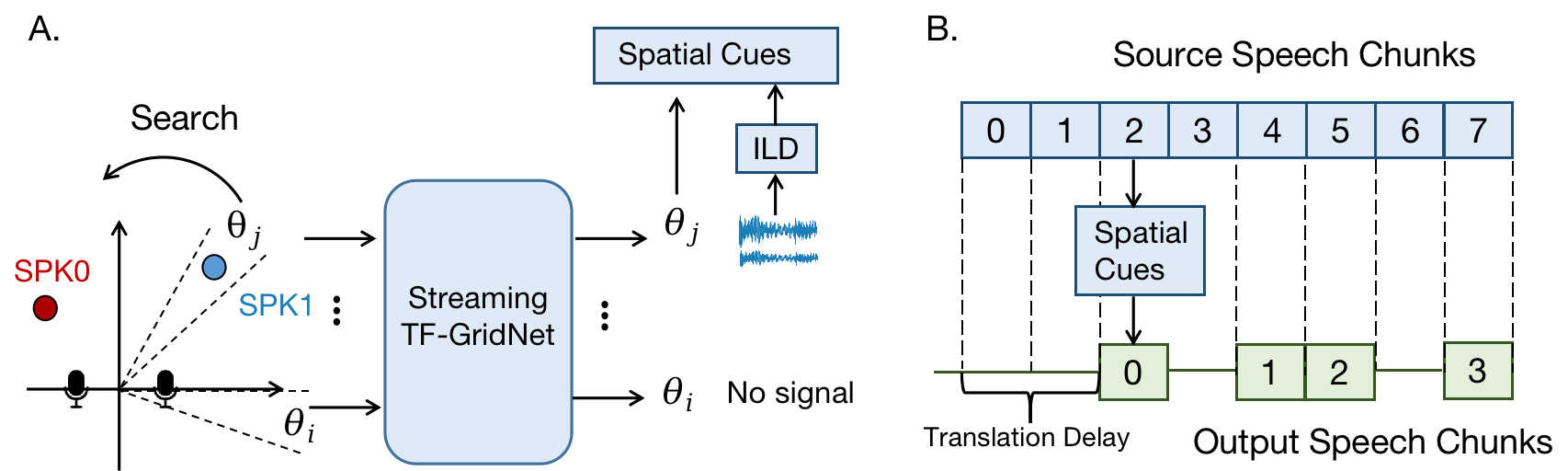}
  \vskip -0.1in
  \caption{{\bf Spatial cues extraction and binaural rendering.}  {(A) shows  search-based joint localization and separation. We divide the space into multiple small angular regions and apply  streaming TF-GridNet on each region. If no source exists (e.g. 
$\theta_i$), the model will output zeros. If a source exists (e.g. SPK1 in $\theta_j$), the model outputs the separated binaural signal (SPK1). The spatial cues are extracted from the estimated angle and ILD of the binaural separated output. (B) shows  binaural rendering in presence of translation latency. The output speech chunk 0 (green block 0) is generated with 2-chunk delay from the source speech chunk 0 (blue block 0). When we render binaural channel output chunk 0, instead of applying the spatial cues of source chunk 0, the spatial cues of current incoming chunk (source chunk 2) is applied. }}
  \vskip -0.15in
  \label{fig:spatial}
\end{figure*}

At a high level, we use binaural headsets with microphones at each ear to perform joint search-based localization and separation. The approach divides the 360-degree angular space into small intervals, and at each angle, a neural network searches for potential speakers. The model is trained to separate and extract a speech source if present or output silence if no speech exists at that angle as shown in Fig.~\ref{fig:spatial}A. There are two key challenges in designing such an algorithm: (1) The model must work in streaming mode for real-time translation, meaning it only processes the current audio chunk without access to future data, unlike non-streaming models that process entire sentences before performing source separation. (2) The model must handle real-world reverberations and Head-Related Transfer Functions (HRTFs). Poorly designed localization and AoA algorithms can be vulnerable to multipath and reverberation. Additionally, varying HRTFs across users can affect the physical separation between left and right microphones. 

\subsubsection{Neural separation algorithm} Formally, say the binaural mixture can be written as,  $x = [x^l, x^r]$, where $x^r \in R^T$ and $x^l \in R^T$ are the audio recordings of length $T$ seconds from the left and right microphones. In noisy environments, the binaural mixture recordings  are composed of speech from $N$ different speakers,  $s^r_i\in R^T$ and $s^l_i\in R^T$, $i \in [0, N)$, as well as  background noise $\widetilde{n}^l \in R^T$ and $\widetilde{n}^r \in R^T$. 
\begin{equation*}
    x^l = \sum_0^{N-1} s^l_i + \widetilde{n}^l, \quad x^r = \sum_0^{N-1} s^r_i + \widetilde{n}^r
\end{equation*}

Say, we define the arrival of angle (AOA) of the $i$th speech source as $\theta_i$. The goal of our joint localization and separation neural network, $\mathcal{M}$,  can be written  as follow.
\begin{equation*}
    \hat{y} = \mathcal{M}(x_l, x_r | \hat{\theta}) = \begin{cases} \
    0 &  \forall i, \theta_i \neq \hat{\theta} \\
    [s^l_i, s^r_i] & \theta_i = \hat{\theta}
    \end{cases}
\end{equation*}

We use the open-source streaming TF-GridNet implementation from~\cite{tsh-chi24} as our backbone  separation model. This model architecture has been demonstrated to be efficient for on-device streaming inference and has also been shown to have a good performance in the presence of real-world reverberation.  To search and extract potential speech from a specific target angle, we shift the binaural input to be aligned at the target angle, $\hat{\theta}$. Specifically, we  shift the right channel of the binaural input by the time difference of arrival (TDoA) corresponding to the target angle $\hat{\theta}$ . Assuming the  sources are in the far-field, our shifted binaural input can be written as,
\begin{equation*}
\begin{split}
    x'_{shift} = [x^l, x^r_{shift} ] = [x^l, x^r(t - TDoA(\hat{\theta}))]
    \end{split}
\end{equation*}
Here, $TDoA(\hat{\theta}) = \frac{d \sin(\hat{\theta})}{c }$ and $d$ represents the distance between the right and left microphones, and $c$ is the speed of sound. At this stage, we assume an average head size of 18 cm for microphone separation and a sound speed of 340 m/s. Note that there are variations in head sizes across real-world wearers, which we account for in our training process as described in~\xref{sec:training_separation}. The key idea is that if a speech source exists at the target angle $\hat{\theta}$, the source signal in the left and right channels will be time-aligned after shifting. The  network  then learns to extract the aligned signal more efficiently, rather than directly conditioning it with the target angle $\hat{\theta}$ \cite{cone-of-silence}.

We apply a Short-Time Fourier Transform (STFT) on the shifted mixture with an STFT window length of 760 samples, or 47.5 ms at 16 kHz, and
a hop length of 640 samples, or 40 ms at 16 kHz. 
Note that the chunk sizes used here are larger than the 8-12~ms chunk sizes used in~\cite{tsh-chi24,semantichearing}. This is because the requirements are different across the two applications.  In~\cite{tsh-chi24,semantichearing}, smaller chunk sizes were needed because the system had to process and play the same audio into the headphones in real time. However, translation requires more context, so larger chunk sizes are used in our implementation. After performing an STFT, we extract the Interaural Phase Difference (IPD) and Interaural Level Difference (ILD) across the binaural channels and concat them with the original STFT spectrogram and feed into the TF-Gridnet Block. We do this for two reasons. First, IPD and ILD features are important for binaural rendering which we describe in more detail in~\xref{sec:binaural_rendering}. Second, IPD/ILD features are derived from domain-specific knowledge and can help the model to more efficiently learn the spatial features in the binaural input. 

\subsubsection{Localization by Clustering}
To process binaural input, we first divide the acoustic  space into $N$ regions, each with $\frac{360}{N}$-degree angular interval. {We use N=36 in our implementation.} We then run the separation model in parallel on these smaller intervals, using the center angle of each interval as the target angle, $\theta_t$. A windowed power threshold 1e-2 is applied with a window size of 0.75 seconds to discard regions with a low probability of containing a  source.

However, in practice, the model may output artifacts as false speech sources even at angles without actual sources. This can happen for two main reasons. First, due to multipath effects, the speech source can falsely appear in adjacent angular intervals as duplicates. Second, indoor reverberation can cause the direct path and strong reflections of a speaker's signal to align at other angular intervals, creating phantom speakers.

We eliminate these false positives by clustering similar separation outputs using the segment-wise similarity algorithm proposed in~\cite{acousticswarm}. Specifically, we first sort all potential speech sources identified by our neural network, ordering them by descending energy after power thresholding. We then iterate through each candidate. For each one, we check if its segment-wise similarity with existing clusters is high. If so, we discard the candidate; if not, we create a new cluster with its separation and target angle. Each final output cluster contains a separated source and a localized angle. 

\subsubsection{Training for Real-world Generalization}\label{sec:training_separation}
Our goal is to generalize to real-world reverberations and multi path as well as variability in HRTFs across wearers. One approach is to collect lots of data with our prototype headset with a large number of wearers in diverse multipath environments. However this is expensive and time consuming. Instead we create synthetic source separation datasets by composing multiple open-source HRTF and multipath datasets. 

Specifically, we use CoVoST2  as our raw speech dataset and WHAM!  as our background noise dataset to create synthetic binaural mixtures. Each training sample in our dataset corresponds to an acoustic scene comprised of 2-3 speech samples from the CoVoST2 dataset as well as  background noise. To convert monaural speech segment from CoVoST2 dataset to binural recording, we convolve each of the speech samples  with room-impulse-response (BRIR). The latter captures
the acoustic transformations caused by Head Related Transfer Function (HRTF) as well as room reverberations. Specifically, we use 4 diferent BRIR datasets: CIPIC~\cite{CIPIC}, RRBRIR~\cite{rrbrir}, ASH-Listening-Set~\cite{ShanonPearce} and CATTRIR~\cite{CATT_RIR}, with a total of 77 different room and wearer configurations. {We split the BRIRs dataset into training, validaiton and testing subsets with no overlap}. For each training sample, we randomly select one BRIR configuration from these 4 datasets. Each BRIR configuration contains a list of channel responses, $h(\theta, \varphi)$, for different azimuth $\theta$ and elevation $\varphi$ angles. Say, $s_0$ and $s_1$ are two monaural speech segments from the CoVoST2 dataset. We randomly sample 2 positions $(\theta_0, \varphi_0)$ and $(\theta_1, \varphi_1)$ and convolute their BRIRs   with two speech segment respectively. With 50\% probability, we add an additional background noise signal, $\widetilde{n}$, sampled from WHAM! dataset to the synthetic mixture $x$. Note that the WHAM! dataset originally have  binaural recordings so  we do not need to again convolute it with BRIRs:
$
    x= s_0 * h(\theta_0, \varphi_0) +  s_1 * h(\theta_1, \varphi_1) + \mathbbm{1}(p < 0.5) \widetilde{n}$, where $p \sim \mathcal{U}(0, 1)$. Totally, we generate 90000 synthetic mixture for training with 50\% probability of adding background noise. For validation and testing set, we respectively generate 3000 samples with background noise and another 3000 without background noise.
We train our separation model with the following training loss function:
\begin{equation*}
\begin{split}
    \mathcal{L}(x, s_0, s_1, \dots s_{N-1}, \hat{\theta}) =  \| \mathcal{M}(x_l, x_r | \hat{\theta}) - \sum_0^{N-1}  s_i \mathbbm{1}(\theta_i \in \hat{\theta} ) \|_1 
    + \\ \lambda \cdot \text{MultiResolution}\Big(\mathcal{M}(x_l, x_r | \hat{\theta}), \sum_0^{N-1}  s_i \mathbbm{1}(\theta_i \in \hat{\theta}) \Big)
\end{split}
\end{equation*}
Here $\|\cdot \|$ is the L1 loss and  MultiResolution$(\cdot)$ is the multi-resolution spectrogram loss. We used 3 resolutions, where we vary the FFT length over $\{1024, 2048, 512\}$, the step size over $\{120, 240, 50\}$, and the window lengths over $\{600, 1200, 240\}$. $\lambda$ is the weighted coefficient we set to 0.1. For the target angle $\hat{\theta}$ for each training sample, with 60\% probability, we randomly select the azimuth that has one speech source and with 40\% probability, we randomly select a azimuth angle  where no source exists.

\subsection{Finetuning for expressive translation}\label{sec:translate:design}
The goal of our translation module includes (1) translate each individual separated source to  speech in the  target language, (2) achieve simultaneous speech translation with real-time on-device inference, and  (3) preserve the vocal and expressive property of speech. As described in~\xref{sec:related}, to the best of our knowledge, there is no open-source model that achieves all these goals. Further, the translation model should be able to operate on the imperfect outputs from the source separation model. Our translation model to achieve this employs a two-pass architecture: we first perform  simultaneous speech-to-text (S2T) translation and then achieve  streaming expressive text-to-speech (T2S) generation.
\begin{figure*}[t!]
\centering
  \includegraphics[width=0.88\textwidth]{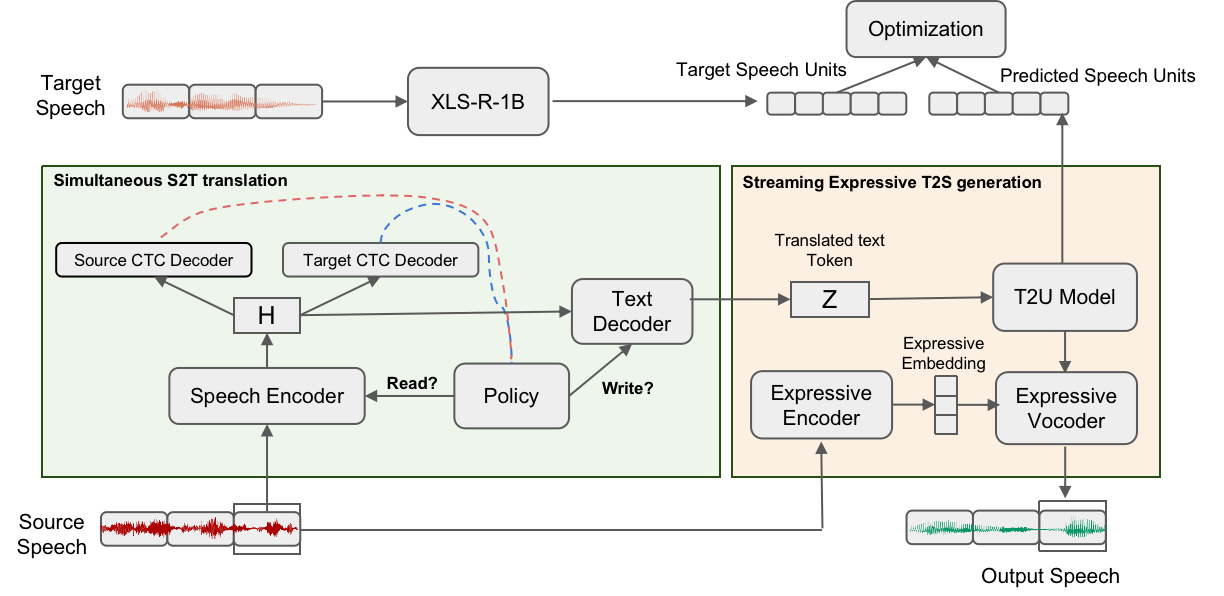}
  \vskip -0.15in
  \caption{{ \bf Simultaneous and expressive speech-to-speech translation.} {(1) In simultaneous speech to text (S2T) translation, a speech encoder  extracts the hidden status $H$ of incoming speech chunks. Source and Target CTC decoders cooperate with a policy algorithm to determine the "WRITE" and "READ" actions for simultaneous translation. When "WRITE" is determined, the text decoder will output target translated text token $Z$. (2) In streaming expressive text-to-speech (T2S) generation, a Text-to-Units (T2U) model converts the text token $Z$ to speech units. The T2U model is trained using  target units extracted from the XLS-R-1B model. Meanwhile, the expressive encoder extracts the expressive embedding from the input speech chunk. Finally, the expressive vocoder takes both predicted units and expressive embedding to re-synthesize the target language.}}
  \vskip -0.15in
  \label{fig:translation}
\end{figure*}

\subsubsection{Simultaneous S2T translation} Our speech-to-text translation module is based on the recently open-sourced StreamSpeech~\cite{streamspeech}, a multi-task simultaneous speech-to-speech translation framework. We compute 80-dimensional mel-filterbank features $X$ from the monaural source speech. These features are then fed into a streaming speech encoder, implemented using the chunk-based Conformer \cite{conformer}, to obtain the hidden embedding $H$ of the source speech.

To simultaneously generate the translated text $Z$ from the hidden embedding $H$, the model learns when it has sufficient context to produce the translation. It does this by learning a policy that decides whether to "READ" the next source state $H$ or "WRITE" the next target text token $Z$. To learn the alignment between the source and target languages, two Connectionist Temporal Classification (CTC) decoders are used: a source text CTC decoder $D^{asr}$ = CTCDecA$(H)$ and a target text CTC decoder $D^{s2tt}$ = CTCDecY$(H)$, both predicting source and target text in a non-autoregressive manner. Additionally, an autoregressive text decoder ARDecY($\cdot$) is employed to use the history of the source hidden state and generate the next translated text token $Z$. While CTCDecA and CTCDecY guide the WRITE/READ policy, the final translated text token $Z$ is generated by the autoregressive ARDecY model when the WRITE action is triggered. This is because autoregressive models yield smoother and more accurate translations than non-autoregressive models~\cite{streamspeech}. The model performs a "WRITE" action only when the CTCDecA model recognizes a new source token and CTCDecY has produced more target tokens than ARDecY. Otherwise, the model continues reading the next hidden state $H$ from the source speech. More details on the policy algorithm can be found in~\cite{streamspeech}.

\subsubsection{Streaming Expressive T2S generation} The next step is to convert the translated text to spoken speech while preserving the vocal and expressive properties of the source speaker. StreamSpeech~\cite{streamspeech} however does not support expressive translation. So instead, we feed the 
translated text token $Z$  to a non-autoregressive text-to-unit (T2U) model and predict the target speech units $U$. The text-to-unit model follows the  T2U architecture in~\cite{t2umodel} and has a T2U encoder and a unit CTC decoder. We also employ an expressive encoder using the ECAPA-TDNN~\cite{seamlessexpressive} architecture  to obtain the expressive embedding $e_p$ from the source speech input. Finally, we apply an expressivity-preserving vocoder~\cite{seamlessexpressive} on the estimated units $U$ to reconstruct the translated speech conditioned on  the extracted expressive embedding $e_p$. 

\subsubsection{Pre-training for simultaneous S2T}
Here, we describe our pre-training pipeline for French-to-English speech-to-text translation. We first train the simultaneous S2T translation module using French-to-English pairs from the CoVoST2 and CVSS-C datasets. Our data pre-processing follows StreamSpeech's pipeline. We select valid French-to-English pairs, resulting in a training set of 207,364 samples of speech-text pairs. The validation and test sets each contain 14,759 pairs, with sample lengths ranging from 3 to 10 seconds. The source speech (French) is resampled to 16 kHz, and we extract 80-dimensional mel-filterbank features, applying global cepstral mean-variance normalization with a 40 ms window. The simultaneous S2T model is optimized via multi-task learning. For source and target text, we use SentencePiece to generate a unigram vocabulary of 6,000 for both languages.

Our training objective $\mathcal{L}_{S2T}$ is  composed of multiple losses including Automatic Speech Recognition (ASR),  Non-Autoregressive Speech-to-Text Translation (NAR-S2TT), and Autoregressive Speech-to-Text Translation (AR-S2TT) tasks, $\mathcal{L}_{S2T} = \mathcal{L}_{asr} + \mathcal{L}_{nar-s2tt} +  \mathcal{L}_{ar-s2tt}$. The ASR task optimizes the output $D^{asr}$ of the source text CTC decoder (CTCDecA) to be close to the source French text. CTC loss is applied between the output token $D^{asr}$ and target French text. The NAR-S2TT task optimizes the non-autoregressive target text CTC decoder (CTCDecY). The CTC loss is applied between the output tokens $D^{s2tt}$ and target translated English text tokens. The AR-S2TT task optimizes the autoregressive text decoder (ARDecY) to generate the target English text tokens. Our loss function is: 
\begin{equation*}
    \mathcal{L}_{ar-s2tt} = -\frac{1}{|Z|} \sum_1^{|Z|} \log (z_i | X_{\leq j}, Z_{< i})
\end{equation*}
Here, $z_i$ is the current output token from ARDecY, $X_{\leq j}$,  the current speech input, and $Z_{< i} = [z_1, z_2 \dots z_{i-1}]$, the history output tokens.  We also apply  multi-chunk training in the same way as \cite{streamspeech} to enable variable latency requirements. The chunk-based conformer in the  speech encoder randomly samples a chunk size $C \sim \mathcal{U}(0, |X|)$.
During training, we use Adam optimizer with an initial learning rate of 1e-3, and inverse-square-root scheduler. We also have a warm-up stage for the first 10,000 steps with the learning rate increasing from 1e-7 to 1e-3. Our training batch size is set to 8.

\subsubsection{Fine-tuning for source separation} The pre-training described above uses datasets that predominantly do not contain interfering speakers or significant noise. However, the output of our separation model, which is fed into the translation model, is imperfect and contains residual distortions from interfering speakers and noise. This may cause performance drop of the translation model, as it has not encountered such distortions during the pre-training procedure. To improve the robustness of the simultaneous S2T model to these distortions and residual interference from source separation, we fine-tune the pre-trained model using the imperfect output of the source separation model. Specifically, we mix 50\% of the data from the original CoVoST2 dataset and 50\% from our separation model output for both the training and validation sets. We use the same multi-task configuration and training objectives as in the pre-training stage. We fine-tune the simultaneous S2T model for an additional 80 epochs with a initial learning rate of 1e-3.

\subsubsection{Training for expressive T2S}
To train the expressive text-to-speech (T2S) module, we use the pretrained PretsselEncoder and PretsselVocoder models from SeamlessExpressive~\cite{seamlessexpressive} as our expressive encoder and vocoder. We froze their parameters during training. We also froze the parameters of the entire simultaneous S2T translation module from the previous training step. In other words, we only updated the parameters of the non-autoregressive T2U model during this training step. To generate the target units to train the T2U model, we applied Meta's large-scale 1 billion-parameter  multilingual pretrained speech model, XLS-R-1B~\cite{babu2021xls}, to the target English speech samples. We extracted the features from the 35th layer of the XLS-R-1B model and mapped features to discrete categories, $\hat{U}$, using  the k-means algorithm (k=10000). We use CTC loss between the target unit, $\hat{U}$, and the estimated unit, $U$, as the training objective. We train the T2U model for another 80 epochs with the initial learning rate set to 1e-3.


In addition to our expressive  T2S model, we also trained a non-expressive  T2S model. For this, we used a pre-trained unit-based HiFi-GAN vocoder to re-synthesize speech~\cite{kong2020hifi}. For target speech, we extracted the discrete units via mHuBERT~\cite{popuri2022enhanced} by applying the k-means algorithm (k=1000) on the representations from the output of the sixth layer. We followed the same training configuration as the expressive model to train the non-expressive T2S model.

\subsection{Binaural rendering of translated speech}\label{sec:binaural_rendering}

The speech translation model described above outputs the monaural speech with expressive preservation for each source in the wearer's acoustic space. Our final step is to render the binaural output for the translated speech.

The main challenge in spatial audio is creating the perception that sound is coming from a specific direction, even when played through headphones~\cite{hrtfuist}. This effect is achieved using spatial cues, which filter sound based on the listener's head, ears, and torso shape (HRTF), as well as the direction of the incoming source speech. By applying these spatial cues, our goal is to simulate the translated speech to match the direction of the source speech. 

Spatial rendering involves two key components: interaural time differences (ITD) and interaural level differences (ILD). Our goal is to preserve both ITD and ILD for binaural audio, which are crucial for human spatial auditory perception~\cite{carlini2024auditory}. The joint localization and speech separation algorithm described in~\xref{sec:jls} provides binaural separated source audio from the mixture, along with the corresponding angle for each speaker. Since this algorithm extracts clean binaural separated source audio, we can estimate the ILD from each speaker and apply it to the translated speech. However, translating ITD across signals without a reference is challenging. Therefore, following prior work~\cite{hrtfuist}, we use a generic HRTF to estimate the ITD corresponding to the source angle. While a generic HRTF does not work well for ILD, as shown in our evaluation section, it provides good results for ITD.

More formally, given the localized angle $\theta_i$ for source $i$, we reference the Generic HRTF dataset, CIPIC~\cite{CIPIC}, and retrieve the corresponding HRTF, denoted as $h(\theta=\theta_i, \varphi=0)$. We convolve this with the monaural translated speech $o_i$ of source $i$ to produce binaural audio samples. We then estimate the ILD features and transfer them from the binaural separated source signal to the binaural translated audio. To preserve ILD features between input and output, we apply an ILD compensation method. Specifically, for source $i$, we first compute its ILD from the separated binaural source speech as  $ILD_i = \|y^r_i\|_1 / \|y^l_i\|_1$. We then scale the translated signal based on the binaural HRTF response and $ILD_i$ as:
\begin{equation*}
\begin{split}
[o^l_i, o^r_i] = [o_i * h^l(\theta=\theta_i, \varphi=0), \\\quad ILD_i \frac{\|h^l(\theta=\theta_i, \varphi=0)\|_1}{\|h^r(\theta=\theta_i, \varphi=0)\|_1}o_i * h^r(\theta=\theta_i, \varphi=0)]
\end{split}
\end{equation*}

We note two key points. First, the above operation can also be performed at each frequency to estimate frequency-dependent ILDs. Additionally, the ITD estimates from the generic HRTF dataset can, in future work, be replaced with personalized HRTFs, which can achieve high-fidelity spatial audio using head scans~\cite{appleheadscan}. 

\noindent{\bf Reconciling translation delays.} An important aspect of merging translation with spatial computing is the delay introduced by the translation model, as shown in Fig.~\ref{fig:spatial}B. The translation model introduces a lag of a few seconds since it requires enough context for accurate translation. This lag is significantly larger than the 40 ms audio chunks processed by the separation and binaural rendering modules. Our key insight is that while translation introduces an inherent delay, spatial awareness for the speaker can still be maintained as long as the translated speech is played from the speaker's current spatial direction. More formally, if the translation delay is $D$ chunks, as shown in Fig.~\ref{fig:spatial}B, the spatial cues from the current source-separated chunk, $i+D$, are applied to the delayed translated chunk, $i$, which is being played through the headphones. 

\begin{table*}[t!]
\vskip -0.15in
    \centering
    \caption{{Runtime measurement decomposition on Apple M2 Silicon. Note that  the AR Text Decoder, T2U Encoder, and T2U Decoder process generated tokens instead of audio chunks. Therefore, we measured runtime based on per-token generation or processing. Our measurements indicated that between 0-10 tokens can be generated or processed for each 960 ms chunk. }}
    \vskip -0.15in
    \begin{tabular}{lccccc}
    \toprule
    Module & Params & Chunk size (ms) & Runtime (ms) & RTF \\
    \midrule
     Separation Model (across all regions) & 640.5K & 40 & 29 & 0.725 \\
     Speech Encoder & 33.45M & 960 & 20.66 & 0.0215\\
    Source Text CTC Decoder (CTCDecA) & 1.542M & 960 & 0.51  &  5e-4\\ 
    Target Text CTC Decoder (CTCDecY) & 1.542M & 960 & 0.51  &  5e-4\\ 
    AR Text Decoder (ARDecY) & 18.84M & 960 & 19.8/token & - \\
    T2U Encoder & 6.3M & 960 & 0.32/token & - \\
    T2U Decoder & 21.94M & 960 & 1.0/token & - \\
    Expressive Encoder/Vocoder & 91.6M & 960 & 63.4 & 0.066 \\
    \bottomrule
    \end{tabular}
   \vskip -0.15in
    \label{table:runtime_comparison}
\end{table*}

\section{Implementation}\label{sec:implementation}
We build our hardware setup around the Sony WH-1000XM4 noise-cancelling headphones. To capture binaural sound, we mount Sonic Presence SP15C microphones onto the outer sides of the earcups. These mics are hooked up to a processing device for processing the audio data. Once the binaural sound is recorded and processed, we replay it through the same noise-cancelling headphones. 

\vskip 0.05in\noindent{\bf Runtime evaluation.} We implement our pipeline on Apple M2 silicon using PyTorch 1.12, which supports the Metal Performance Shaders (MPS) backend for Apple silicon. The "torch.mps" library is used to run model inference and measure runtime on the Apple M2. We decompose each component of our pipeline and measure the runtime for each module, as shown in Table~\ref{table:runtime_comparison}. The separation model processes 40 ms audio chunks, while the translation model runs every 960 ms. For accurate runtime measurements, we warm up the models by running them for 10 chunks, then measure the average runtime over 200 chunks. For each chunk, we start a timer before feeding it into the model, run "torch.mps.synchronize()" to ensure all kernels in all streams on the MPS device have completed, and record the runtime.

The table shows that Apple M2 silicon efficiently runs our models. It also reports the run time factor (RTF), which is the fraction of the chunk time needed to process it. An RTF of less than one indicates real-time operation. Despite the translation models having significantly more parameters than the separation model, they run more efficiently on the M2, likely because Apple silicon is optimized for transformer neural network architectures, which are core to our translation models. In contrast, the separation model uses LSTMs and other components, which do not seem to be optimized on the Apple silicon. However, we note that the RTF considering the cumulative runtime  across all the models is still less than 1. 


\section{Real-world evaluation}


\begin{figure*}[t!]
\centering
  \includegraphics[width=1\textwidth]{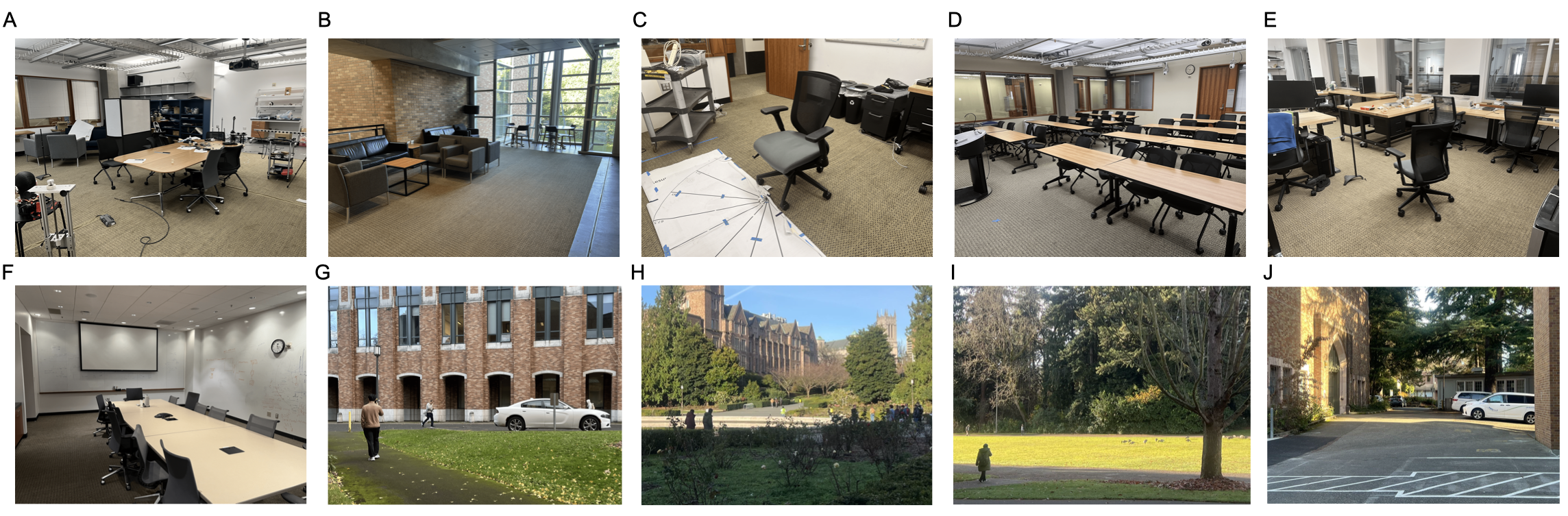}
  \vskip -0.1in
  \caption{{\bf Real-world evaluation settings.} A-J show ten different unseen multipath environments tested in our real-world generalization evaluation. A-F shows indoor spaces including office spaces, class rooms, common open spaces, conference rooms and as well as work spaces. {G-J shows outdoor spaces like school area, fountain park, public picnic lawn and parking lot.} }
  \vskip -0.15in
  \label{fig:rooms}
\end{figure*}

{To evaluate the  performance of our translation system, we tested it in previously unseen real-world environments with participants not included in the training data. We recruited 10 participants (4 female, 6 male), aged 20–35 years, comprising 4 native and 6 non-native English speakers. These 10 participants attends the entire end-to-end evaluation process involved: (1) collecting data in unseen real-world environments, (2) running the algorithmic pipeline end-to-end to generate binaural translated English speech, and (3) playing back the translated speech for participants to assess its accuracy, quality, and preservation of spatial cues.}


{We conducted real-world experiments in 10 distinct, previously unseen indoor environments (e.g., office, living room, conference room, classroom) and 4 outdoor settings (e.g., fountain garden, parking lot, picnic area, school area), as shown in Fig.\ref{fig:rooms}. Indoor environments featured background noises such as electrical hums, air conditioning, and footsteps, while outdoor settings included dynamic, uncontrollable sounds such as birds, human chatter, wind, airplane and traffic.} In each environment, participants wore our headset hardware, which was wired to capture high-quality binaural recordings at 16kHz. Loudspeakers were positioned at different angles and played French speech to simulate human speakers in these settings. A whiteboard with angle markings (Fig.~\ref{fig:rooms}C) was used to record the ground-truth angles of the loudspeaker placements.


For each participant, we selected two loudspeakers from the following options: Korono Bass+ Mini Speaker, IK Multimedia iLoud Micro Monitor, and SRS-XB10. One loudspeaker was connected to a laptop via wire, while the other was connected via Bluetooth, enabling us to use the "sounddevice" library to play audio through both speakers simultaneously. For each mixture, we randomly sampled two French audio clips from the test set of the CoVoST2 dataset and played them concurrently through the speakers. {The amplitude of each French source was independently randomized between 0.2 and 1.0. Consequently, in the collected mixture,  the power difference between two audio sources at the headset ranged from -15 dB to 15 dB.}  Each participant contributed 7 to 8 real-world audio mixtures, with each mixture lasting between 5 and 15 seconds.  {Finally, we also conducted a  varying-distance evaluation. We collected another 27 real-world samples, where the distance between the wearer and each loudspeaker independently varied between 0.75 m and 2.5 m.} All recorded mixtures were processed through our complete speech translation pipeline, which included separation, translation, and rendering, to produce binaural translated English speech.


\subsection{Subjective evaluation}
{We played back the binaural translated English speech from real-world recordings to the headphones of 10 participants. We then conducted a subjective evaluation of the end-to-end system to address the following research questions from a user perspective: 
(1) {\it How accurately is the translation's meaning preserved?} (2) {\it How well are speaker characteristics preserved after translation?} (3) {\it How accurately are spatial cues preserved?} (4) {\it What are the subjective reactions to different translation latencies?} (5) {\it What are subjective reactions to varying noise cancellation levels?} and (6) {\it What is the overall qualitative feedback on our end-to-end translation system?} }


\subsubsection{Human evaluation of end-to-end binaural source separation and translation.} \label{sec:rel}
We designed a listening survey where participants rated our translation system's output. {In addition to the 10 participants from our in-the-wild data collection, we recruited 19 more participants for  this listening survey. Participants were recruited both from within and outside our institution, encompassing the broader metropolitan area. The only inclusion criteria were that participants must be adults and must not have medically diagnosed hearing loss. In total, we had 29 participants, comprising 14 females and 15 males, with ages ranging from 19 to 53. The demographics include  6 caucasians, 14 asians and 9 participants of Indian sub-continent descent. Approximately 35\% of participants were native English speakers, and 80\% were bilingual. Around 73\% of the participants had a technical background. The study included 12 sections to evaluate translation quality in indoor experiments and 6 sections for outdoor experiments.
} In each section, participants first listened to a 6 to 10 seconds French audio sample and noted the speaker characteristics. We also provided the corresponding English translation. Participants then listened to three distinct audio clips: 1) the output of the StreamSpeech~\cite{streamspeech} model using the raw recording as input, without any source separation algorithms, i.e., without any spatial awareness, 2) the output of our model with source separation but without expressive embeddings, and 3) the output of our model with both source separation and expressive embeddings. The three clips were presented in random order in each section.

After listening to the  original French  sample and outputs from the three models in a random order, we ask the participants to rate the translation quality by asking them the following questions~{\cite{seamlessexpressive}}:
\begin{enumerate}
    \item \textbf{Semantic consistency}: \textit{In terms of the MEANING in the translated speech, how similar was the sample in comparison to the true translation? 1 - Very different, 2 - Some similarities, but more differences, 3 - Some differences, but more similarities, 4 - Very similar}
    \item \textbf{Speaker similarity}: \textit{In terms of SPEAKER CHARACTERISTICS, how similar was the sample in comparison to the original French speaker? 1 - Very different, 2 - Some similarities, but more differences, 3 - Some differences, but more similarities, 4 - Very similar}
\end{enumerate}


\begin{figure*}[t!]
\centering
\includegraphics[width=1.7\columnwidth]{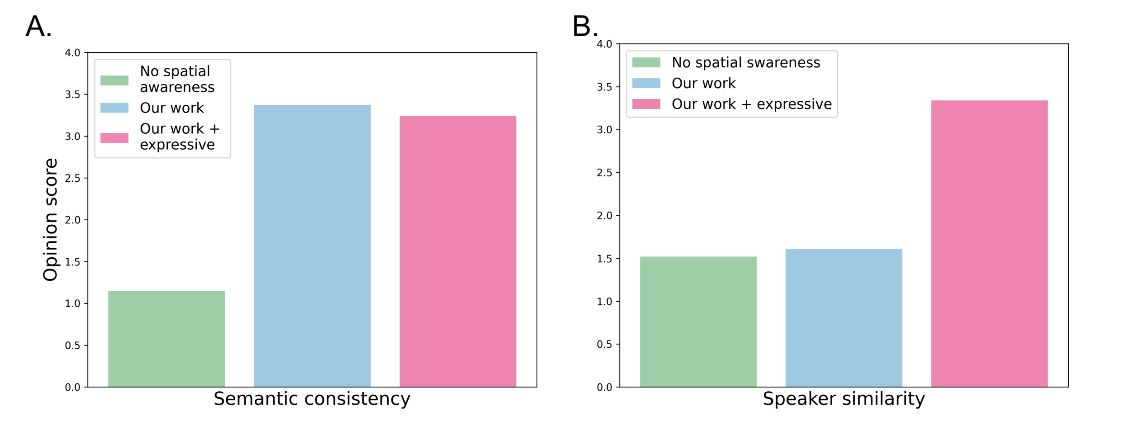}
\vskip -0.1in
\caption{{\bf Subjective evaluation of semantic consistency and speaker similarity.}  The left figure shows the mean opinion
score for existing translation models without any spatial awareness, our work which performs binaural source separation and translation and finally our work with expressive translation. The right figure shows the corresponding results for speaker similarity between the original French speech and the generated English translation (we use a 1-4 scale).}
\vskip -0.2in
\label{fig:mos}
\end{figure*}




{Fig.~\ref{fig:mos} shows that without spatial awareness—i.e., without our binaural source separation module—the mean opinion scores for both semantic consistency and speaker similarity were low, at 1.149 and 1.518, respectively. This is expected, as translation models assume the input corresponds to a single speaker, resulting in garbled translations when multiple interfering speakers are present. With binaural source separation, our model achieves semantic consistency scores of 3.372 and 3.241 for non-expressive and expressive models, respectively. The expressive model has slightly lower semantic consistency due to incorporating speaker characteristics, which includes some noise in the real-world, slightly degrading translation performance. As expected, adding expressive embeddings to our model improves speaker similarity from 1.818 to 3.455.}

{To investigate user perceptions and preferences between non-expressive and expressive translations in our target multi-speaker conversation setting, we included six additional sections in the survey. Each section featured a multi-speaker conversation between two French speakers. Participants were provided with two translated samples generated by our system:  the corresponding English translations of the conversation (1)  without expressive embeddings and (2)  with expressive embeddings. After listening, participants rated the semantic consistency and speaker similarity of the translations, as in prior sections. Additionally, they answered a question about their preference between expressive and non-expressive translations in the multi-speaker conversation setting. The consistency scores were 3.64 for the non-expressive model and 3.28 for the expressive model. The speaker similarity scores were 1.81 for the non-expressive model and 3.46 for the expressive model.}

{Notably, 73.3\% of the samples were preferred with expressive embeddings, highlighting the importance of preserving speaker characteristics in multi-speaker translation. Participants were also asked to provide qualitative feedback on expressive and non-expressive translations. Most participants indicated a preference for expressive translations, citing the ability to differentiate speakers in the conversation as a key factor. Three participants noted that the non-expressive translations sounded robotic. However, two participants remarked that in some of the samples, the expressive translation preserved  accents and noise present in the original French speech, leading them to prefer the clearer non-expressive version for those specific samples. A participant  noted that they might have changed their preference to expressive translation for some of the samples, if they heard a longer duration conversation with multiple turns. Their reasoning was that  they needed to hear more than a single sentence to get familiar with the speaker characteristics and accent.
}

\begin{figure}[t!]
\centering
\includegraphics[width=\columnwidth]{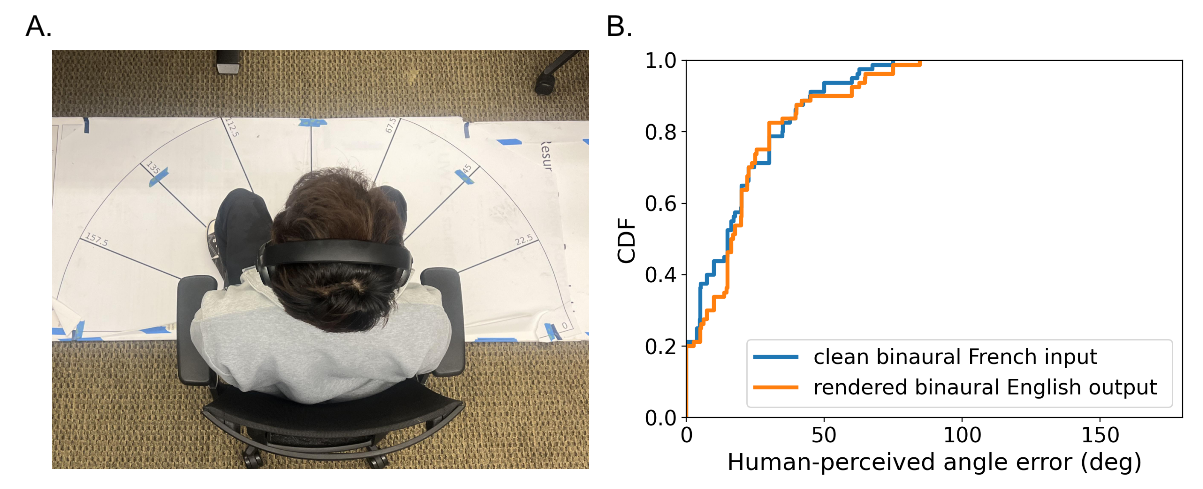}
\vskip -0.2in
\caption{{{\bf Real-world evaluation for user-perceived spatial cues {(10 participants)}.} (A) Experiment setup for the human auditory localization task with the virtual auditory display in the headphone. (B) CDF of the angular errors between the ground truth source directions and the users’ perceived directions obtained for both the clean source French speech as well as the binaural rendered English speech output. }}
\vskip -0.2in
\label{fig:real_angle}
\end{figure}

\subsubsection{Evaluating user-perceived spatial cues} To evaluate the spatial preservation performance of our system in real-world, we conducted human-perception experiments with {the 10 participants.} Following~\cite{hrtfuist}, we conducted a human auditory localization task using a virtual auditory display to evaluate the effectiveness of spatial cue preservation with our real-world audio. For each individual French speech in our real-world recorded audio clip, we prepare 2 types of sources (1) raw single-speaker binaural French audio additionally recorded at the same angle in the mixture (2) its corresponding translated binaural English speech after separation, translation, and rendering. The more accurately spatial cues are preserved, the more closely humans perceive the direction of the translated English speech to match that of the  French speech.

\begin{figure*}[t!]
\centering
\includegraphics[width=1.9\columnwidth]{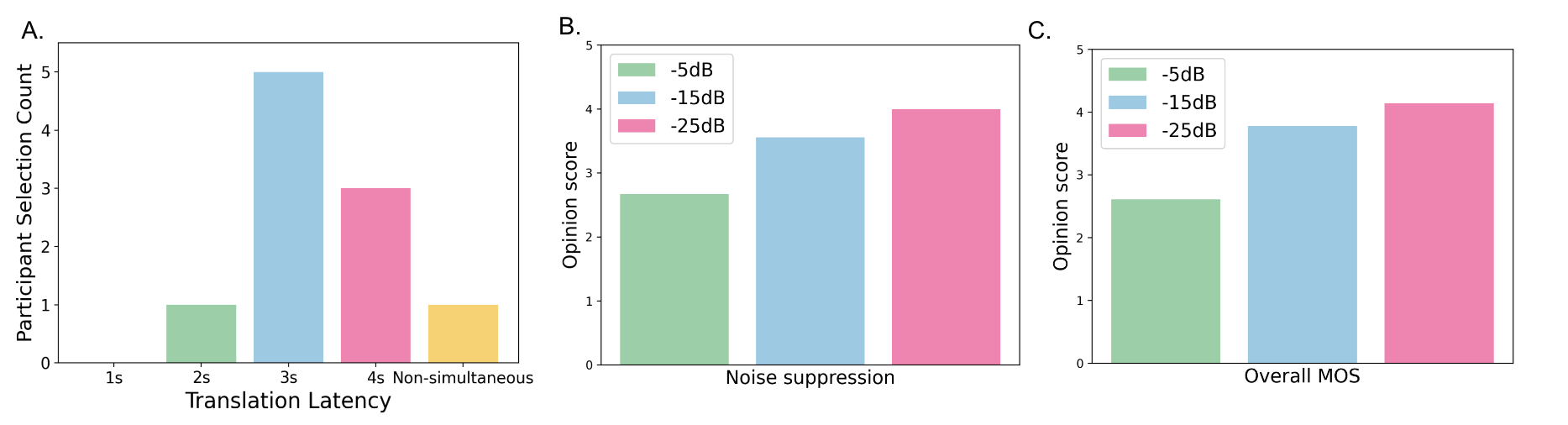}
\vskip -0.1in
\caption{{\bf Subjective latency and noise cancellation evaluation.} {(A) shows the preference for different translation latencies  across 10 participants.} (B) shows the mean opinion MOS score for the noise suppression quality reported for different noise cancellation levels. (C) shows the corresponding overall reported mean opinion score (MOS).}
\vskip -0.2in
\label{fig:cancelation}
\end{figure*}

In the evaluation, each participant wore a headset and sat on a chair placed on a whiteboard with angle markings, as shown in Fig.~\ref{fig:real_angle}A. We played 16 raw French binaural speech samples and the corresponding rendered English speech through their headphones. The 16 samples were played in random order to prevent participants from predicting the direction based on adjacent samples. Additionally, we played a 1-second burst of diffused white noise before each sample. After listening, participants were asked to predict the sound source direction by rotating their body toward the perceived direction. We compared the errors between the ground truth source directions and the participants’ perceived directions for both the raw French speech and the translated English speech output by our system. { As illustrated in Fig.~\ref{fig:real_angle}B, the median perceived angular error for the rendered English speech increased slightly from 15.0 degrees to 16.7 degrees compared to the original French speech.} This shows that our   method preserves the spatial cues of the source speech and has minimal impact on participants' perception of  direction compared to the untranslated speech.

\subsubsection{{Translation delay subjective evaluation.}}\label{sec:delay} {To  explore user perceptions and reactions to different translation latencies, we conducted an additional user evaluation. Specifically, we adjusted the chunk size of the conformer speech encoder to 160 ms, 320 ms, 960 ms, 1920 ms, and a non-simultaneous mode to run our translation pipeline on recorded real-world French mixtures. The corresponding translation latencies for chunk sizes of 160 ms, 320 ms, 960 ms, and 1920 ms were approximately 1 s, 2 s, 3 s, and 4 s, respectively. In the non-simultaneous mode, the conformer encoder began translating only after the entire French speech input was processed, i.e., at the end of the speaker's turn.}

{We conducted the translation latency evaluation with 10 participants. Each participant was seated in front of two loudspeakers while wearing our headset. One loudspeaker played the raw French speech, and the other emitted background chatter noise to simulate a noisy environment. Participants were instructed to focus on the target French loudspeaker while we played the translated English speech with varying latencies. During translation playback, the active noise cancellation (ANC) feature of the headphones was also enabled. To give participants a sense of translation accuracy, we provided the reference English text for comparison. After listening, participants were asked to select their preferred translation latency, considering both accuracy and delay.}

{As shown in Fig.~\ref{fig:cancelation}A, five participants preferred a latency of 3 seconds, three preferred 4 seconds, one preferred 2 seconds, and one preferred the non-simultaneous mode. Participants were also asked to provide qualitative feedback on the different latencies. Most mentioned that the 1-second latency often resulted in half sentences being translated incorrectly, while the 2-second latency caused errors in one or two words. Considering the tradeoff between latency and translation accuracy, the majority preferred a latency of 3 or 4 seconds. The participant who preferred a 2-second latency explained that they could tolerate minor word errors and valued smoother, more seamless conversations. Meanwhile, the participant who favored the non-simultaneous mode prioritized accuracy and found it acceptable to translate one sentence at a time. Finally, we provide a quantitative analysis of the trade-off between latency and translation accuracy in Fig~\ref{fig:al_bleu}.}

\subsubsection{Subjective noise cancellation evaluation.} The translated speech must be played into the wearer's ears while they still hear the original speakers in the environment. As shown in Fig.~\ref{fig:teaser}(C), our system activates noise cancellation on the headsets to suppress the original speakers, allowing the wearer to focus on the translated version. Here, we aim to evaluate: {\it How much noise cancellation is needed for real-time translation on a headset?}

Specifically, the headset suppresses the French speaker's voice along with any interfering speech and plays the corresponding translated English audio. To investigate this, we asked the 18 participants in our listening study to evaluate this question. Each participant listened to two sets of samples, each featuring a combination of the original French speaker and the translated speech. The French speaker's voice was attenuated at three levels: 5 dB, 15 dB, and 25 dB, and participants evaluated each combination.

After listening to audio samples at the three suppression levels, we asked the participants to rate the audio samples by asking them the following questions:

\begin{enumerate}
    \item \textbf{Noise suppression}: \textit{How INTRUSIVE/NOTICEABLE were the
INTERFERING SPEAKERS and BACKGROUND NOISES?  1 - Bad, 2 - Poor, 3 - Fair, 4 - Good, 5 - Excellent}
        \item \textbf{Overall MOS}: \textit{If the goal is to focus on the  translated speech,
how was your OVERALL experience?  1 - Bad, 2 - Poor, 3 - Fair, 4 - Good, 5 - Excellent}

\end{enumerate}

Fig.~\ref{fig:cancelation}B, C show that as the attenuation of the French speakers increased, participants gave higher opinion scores for both noise suppression and the overall mean opinion score (MOS). There was a notable jump in scores between 5 dB and 15 dB, with a smaller increase when the attenuation reached 25 dB. These results highlight that participants preferred greater attenuation of the original French speaker to better understand the English translation, indicating that using noise cancellation  is crucial for effective speech translation on hearables. Further, attenuation levels between 15 and 25 dB were sufficient to achieve reasonable MOS values. 


\subsubsection{{Additional Human Feedback.}} 
{In addition to collecting human feedback on expressive translation (\xref{sec:rel}) and translation latency (\xref{sec:delay}), we asked three additional subjective questions to better understand the participants' overall experience with our translation systems:  (1) {\it Where do you see yourself using such a
system?} All participants mentioned using the translation system while traveling abroad in public spaces, such as museums, streets, bars, and restaurants. Additionally, three participants highlighted its usefulness for socializing with international attendees at conferences. One participant noted that existing translation apps often fail in noisy environments or when multiple people are speaking simultaneously, emphasizing the need for a system capable of functioning effectively in crowded settings. Another participant shared that they feel more confident speaking in their native language but struggle to find the right words in English. They noted the potential of such a tool to facilitate communication in real-world situations, enhancing their confidence. (2) {\it Would you prefer the speech translation running on the device or in the cloud?}  Six participants expressed a preference for on-device translation due to concerns about latency and privacy. Three participants indicated that either option would be acceptable, while one participant preferred cloud-based translation, citing higher accuracy enabled by larger models running in the cloud.
(3) {\it What can be further improved in our current translation systems?} Three participants suggested that the quality and naturalness of expressive speech generation could be improved. One participant proposed adding a notification feature to indicate when translation playback is about to start, allowing users to prepare for it. Two participants recommended making the system more customizable and adaptive for different scenarios—for instance, simultaneous speech translation for casual, social, and informal interactions, and non-simultaneous but highly accurate translation for high-stakes scenarios, such as those involving health and safety. }

\subsection{{Metric evaluation on real-world data}}

{We evaluate the quantitative metrics typically used in translation, speech separation, and localization literature on our real-world collected data. Our entire speech translation pipeline is tested in an end-to-end manner using real recordings of concurrent speakers. First, the recordings are processed through our source separation module, and the localization accuracy is reported in \xref{sec:loceval}. For the speech translation module, we compare four different model configurations: (1) Non-finetuned S2T with non-Expressive T2S (baseline from StreamSpeech\cite{streamspeech}), (2) Non-finetuned S2T with Expressive T2S, (3) Finetuned S2T with non-Expressive T2S, and (4) Finetuned S2T with Expressive T2S. The chunk size for all translation models is set to 960~ms. } 

\subsubsection{Localization evaluation}\label{sec:loceval}

\begin{figure}[t!]
\centering
\includegraphics[width=\columnwidth]{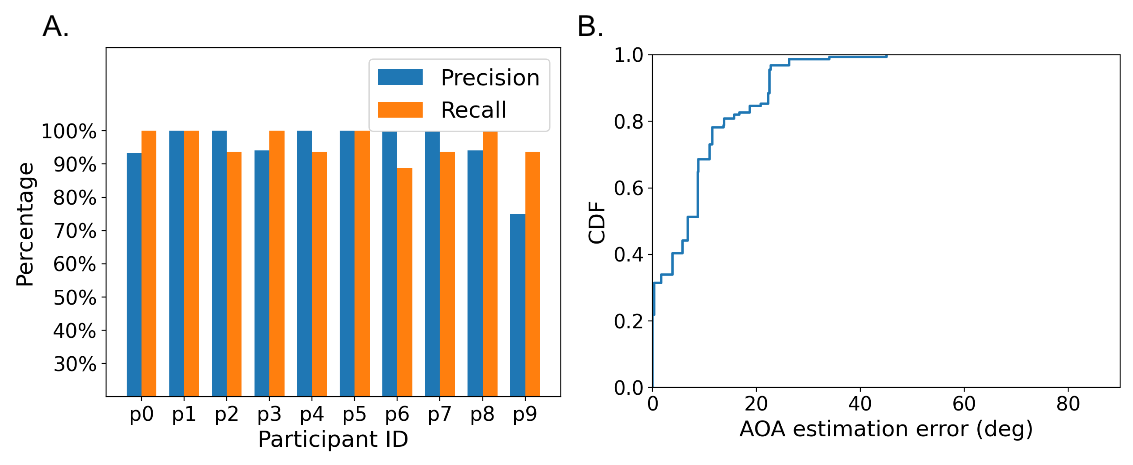}
\vskip -0.2in
\caption{{{\bf Real-world evaluation for joint separation and localization.} (A) Precision and recall  for each participant. (B) The CDF curve of AOA estimation error for each separated sound source obtained from the real-world mixture. } }
\vskip -0.2in
\label{fig:loc}
\end{figure}

First, we evaluated the performance of our localization and separation module in real-world scenarios. Fig.~\ref{fig:loc}A shows the precision and recall for identifying the two interfering French speakers across ten participants, where P0-P5 are indoor and P6-P9 are outdoor. For indoor scenarios, the mean precision and recall across 6 participants were 97.08\% and 97.92\%, respectively. {For outdoor scenarios, the mean precision and recall across 4 participants were 92.3\% and 94.1\%. The results demonstrate our algorithm accurately identifies the number of speakers in real-world conditions. }
Fig.~\ref{fig:loc}B shows the cumulative distribution function (CDF) of errors  in angle-of-arrival (AoA) estimates across all French mixtures collected from 10 participants. { The plots indicate that the median AoA error was 6.8 degrees, with a 90th percentile error of 22.5 degrees for the two concurrent speakers. The indoor median AoA error was 3.8 degrees, while the outdoor median AoA error increased to 8.8 degrees. In the  varying-distance experiments, the precision and recall across three participants were 96\% and 94.2\%, respectively. The corresponding median AoA error was 3.8 degrees,  with a 90th percentile error of 11.5 degrees. }This demonstrates that the separation module effectively estimates and localizes sound sources in unseen real-world reverberant and noisy outdoor environments, as well as at varying distances. Moreover, our system successfully generalized to participants and corresponding HRTFs that were not part of the training data.

\subsubsection{Translation Metrics Evaluation}

We evaluate the following  metrics to measure both the semantics and the expressive aspects of our joint source separation and streaming translation models.

\begin{enumerate}


    \item \textbf{ASR-BLEU}: BiLingual Evaluation Understudy (BLEU) is a  commonly used metric for evaluating the quality of text generated by machine translation. Automatic speech recognition (ASR) models convert speech into text. BLEU scores can be calculated after applying ASR to the translated speech generated by our model. This metric evaluates how well expressive speech-to-speech models maintain translation quality~\cite{seamlessexpressive}. 
\item \textbf{AL}: Average lagging (AL) for speech-to-speech translation models quantify the degree the user is out of sync
with the speaker, in terms of the amount of time~\cite{streamspeech}.
\item \textbf{AutoPCP score}: PCP  is a measure used to evaluate prosodic preservation in expressive translation systems, based on human judgments  of how similar two spoken utterances sound in terms of prosody. AutoPCP is a  network trained to predict PCP scores for sentence-level prosody similarity~\cite{seamlessexpressive}.

\item \textbf{Vocal style similarity (VSim)}: This metric measures the similarity between source speech and translated speech~\cite{seamlessexpressive}. We extract embeddings from both the generated and source speech using a pretrained WavLM-based encoder~\cite{Chen_2022}, and then measure the cosine similarity of the embeddings.
\end{enumerate}
We use   SimulEval~\cite{simuleval2020}  to  measure ASR-BLEU and AL, and  "stopes"\cite{stopes}  to evaluate  Auto-PCP and VSim.

\begin{table*}[t!]
    \centering
    \caption{Results on real-world collected data for Fr-En translation module. (6 indoor environments {and 4 outdoor environments).}}
    \vskip -0.15in
    \begin{tabular}{lcccccc}
    \toprule
    Speech Input  &  Finetuning  & T2S    & ASR-BLEU & AL(s) & Auto-PCP & VSim \\
     &  for separation      &  &  &  & &\\
    \midrule
     real-world data  &  No & non-Expressive & 18.06  & 3.42 &  2.06 & 0.038 \\ 
            &  No & Expressive & 16.50  & 3.41 & 2.25 & 0.233 \\ 
           &       Yes & non-Expressive  & 22.07 & 3.42 & 1.89 & 0.013 \\ 
           &       Yes & Expressive& 18.54  & 3.43 & 2.30 & 0.250 \\ 

    \bottomrule
    \end{tabular}
    \vskip -0.15in
    \label{translation-benchmark}
\end{table*}

\begin{figure}[t!]
\centering
\includegraphics[width=0.8\columnwidth]{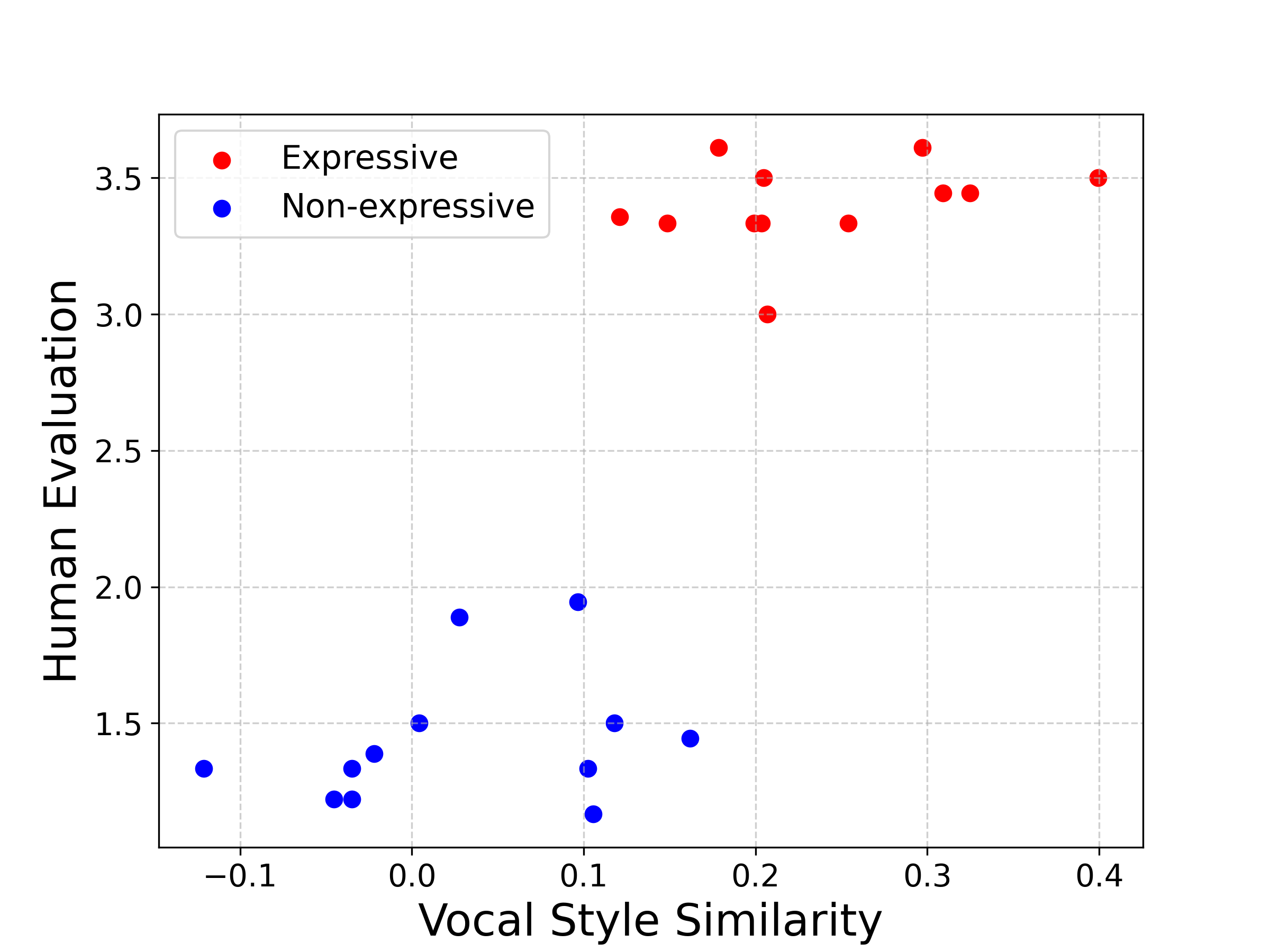}
\vskip -0.15in
\caption{{\bf {Contextualization of the  VSim metric.}} {For each sample in our listening survey, X-axis is the  Vsim values and y-axis is the corresponding human-evaluated speaker similarity score averaged across all participants. It shows a strong correlation between VSim  and participant preferences for speaker similarity. }}
\vskip -0.2in
\label{fig:vsim}
\end{figure}

\begin{figure}[t!]
\centering
\includegraphics[width=0.7\columnwidth]{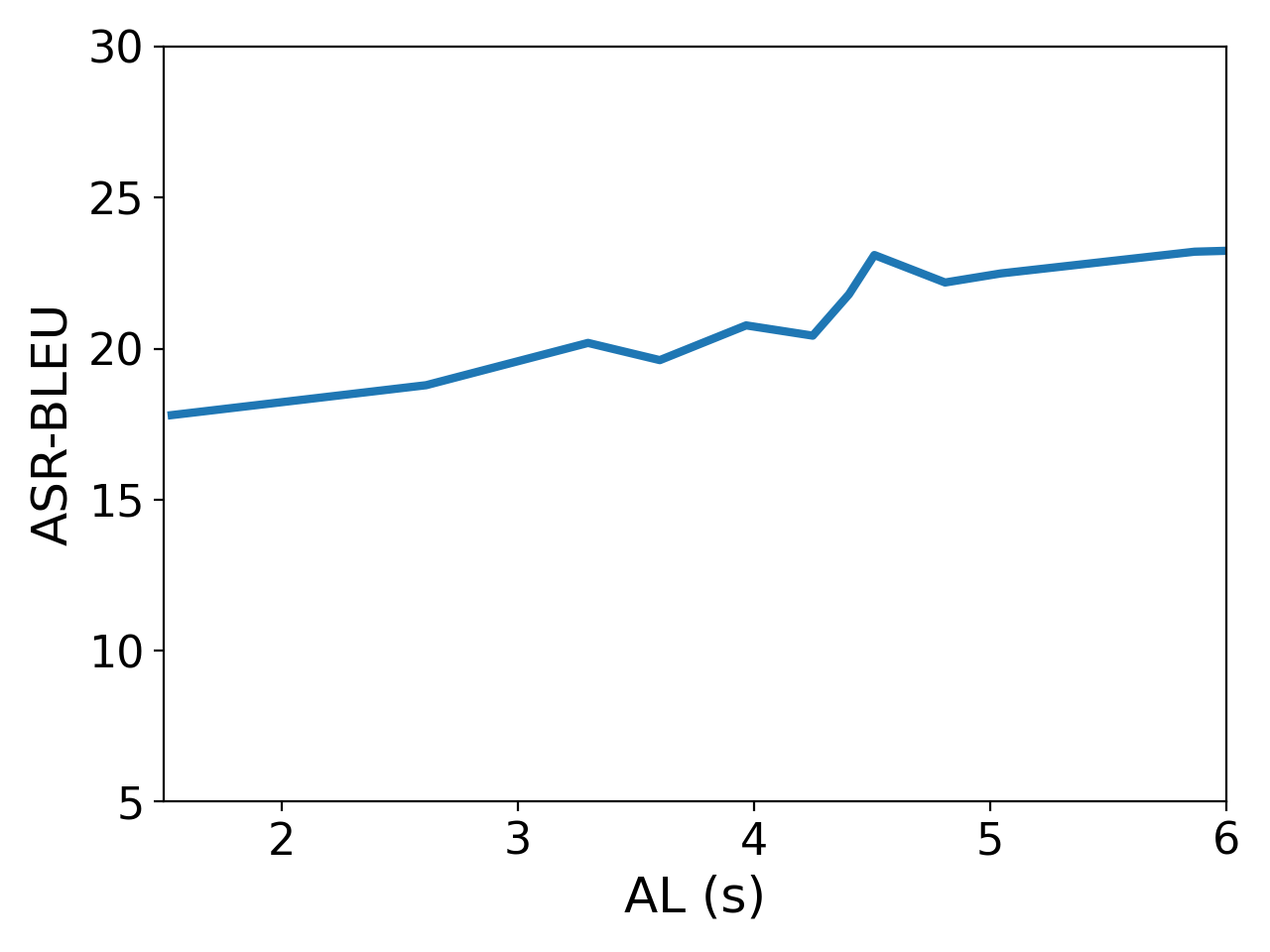}
\vskip -0.15in
\caption{{\bf Tradeoff between ASR-BLEU and AL.} {We ran our model with different input  chunk sizes and measured AL as well as ASR-BLEU. The figure shows that the larger chunk sizes lead to higher latencies and larger ASR-BLEU values. }}
\vskip -0.2in
\label{fig:al_bleu}
\end{figure}


\vskip 0.05in\noindent{\bf Results.}{  The translation results   are shown in Table.~\ref{translation-benchmark}. Our finetuning procedure  increases ASR-BLEU by 4 points for our non-expressive model and by 2 points for our expressive model. Compared to the non-expressive model, the expressive model achieved a higher VSim score, increasing from 0.013 to 0.25, and improved Auto-PCP from 1.89 to 2.30. Across the 10 environments, the indoor ASR-BLEU score was 16.89 with an average lagging (AL) of 3.26s, {while the outdoor ASR-BLEU score was 20.8 with an AL of 3.68s. This difference may be because the translation model automatically waits for more audio samples in outdoor scenarios to confidently generate translation outputs, which in turn results in a higher ASR-BLEU score. Additionally, the outdoor VSim score (0.265) was higher than the indoor VSim score (0.24). These results demonstrate that our spatial translation pipeline remains robust in outdoor scenarios with dynamic noise.}} Compared with Finetuned and Non-expressive model, the Finetuned and Expressive model have a lower ASR-BLEU. This is because our expressive encoder is frozen, and we found that the distortion and noise of input speech (even raw French speech in CoVoST2 has some noise and distortion) will also be preserved in the generated translated speech. Especially, in the real-world experiment, the recorded raw French CoVoST2 dataset was played and recorded again by loudspeaker and microphone, which led to  distortion and muffling of the  speech.

{To better contextualize the VSim metrics, where we see the most improvement,  we plot the correlation between VSim  and human-evaluated speaker similarity scores for the translated samples from the listening survey (see Fig.~\ref{fig:vsim}) . The results show a strong correlation between Vsim and human-evaluated scores, indicating that the Vsim metric effectively reflects human opinions.}

{We also measured the translation metrics for our varying-distance experiment (0.75m-2.5m). The Finetuned Non-Expressive model achieved an ASR-BLEU score of 18.70 with an AL of 3.3s, while the Finetuned Expressive model achieved 16.94 with an AL of 3.4s. For expressive preservation, the Finetuned and Expressive model achieved a VSim score of 0.249 and an Auto-PCP score of 2.20.  These results demonstrate that our translation pipeline remains effective even when sound sources are at varying distances.}




{Finally, we explored the trade-off between the translation quality and  latency on  real-world samples. We ran  the Finetuned and Non-expressive model with 11 different input audio chunk size: 320ms, 640ms, 960ms, 1280ms, 1600ms, 1920ms, 2240ms, 2560ms, 2880ms, 3200ms, and 4800ms, resulting in different average laggings (ALs). Fig.~\ref{fig:al_bleu} shows that as expected higher translation latency leads to higher ASR-BLEU values.}


\begin{table*}[t!]
    \centering
    \caption{{Joint separation and localization benchmark results. (BG - background noise)}}
    \vskip -0.15in
    \begin{tabular}{lccccc}
    \toprule
    Dataset & Precision & Recall & SI-SDRi(dB) & Angular error (degrees) \\
    \midrule
     Separation w/o BG & 0.966 & 0.99 & 14.52 & 4.13 \\
     Separation w BG  & 0.952 & 0.97 & 10.70 & 4.12 \\
    \bottomrule
    \end{tabular}
    \label{table:separation_benchmark}
\end{table*}

\section{Benchmarking the models}\label{sec:benchmarks}
While real-world human evaluation with our wearables is important to demonstrate generalization to unseen environments and wearers, for completeness,  we benchmark our models using test datasets.

\subsection{Benchmarking  localization and separation}
We first evaluate our joint localization and separation pipeline on a test set consisting of 3,000 synthetic mixtures with background noise and 3,000 without, generated using the datasets described in~\xref{sec:training_separation}. The input SI-SDR for the test set without background noise has  0.1 dB mean value and 10.5 dB standard deviation, while for the test set with background noise, its  mean value is -1.8 dB  with 7.7 dB standard deviation.  If the model mistakenly outputs an additional speech source, it is considered a false positive, and if it fails to recognize a  speech source, it is a false negative.  As shown in Table~\ref{table:separation_benchmark}, our system achieved 95.3\% precision and 99.0\% recall without background noise, and 95.2\% precision and 97.0\% recall with background noise. The average angle of arrival (AoA) estimation error was similar in both cases. AoAs were computed for both French speakers in the mixture, and averages were taken across the speakers and the test set.

 To evaluate separation quality, we computed the Scale-Invariant Signal-to-Distortion Ratio improvement (SI-SDRi)~\cite{tsh-chi24} for both separated speakers. Our separation model achieved an average SI-SDRi of 14.52 dB with two speakers and 10.79 dB with two speakers plus additional background noise. The loss in SI-SDR improvement with background noise is likely due to our separation model's limited size of 640.5K parameters, which must handle both source separation and noise suppression. This is significantly smaller than the 2M parameters used in~\cite{tsh-chi24}. Increasing the number of parameters has been shown to enhance the neural network's ability to perform both tasks effectively~\cite{acousticswarm,tsh-chi24}.

\begin{table*}[t!]
    \centering
    \caption{Benchmarking results  for Fr-En translation module. (BG - background noise)}
    \vskip -0.15in
    \begin{tabular}{lcccccc}
    \toprule
    Speech Input  &  Finetune  & T2S    & ASR-BLEU & AL (s) & Auto-PCP & VSim \\
      &  for Separation     &  &  &  & &\\
    \midrule
     Raw single  & No & non-Expressive & 19.54 & 2.81 & 2.097 & 0.0174\\ 
     speaker speech      &   No & Expressive & 17.32  & 2.81 & 2.375  & 0.2526\\ 
           &      Yes & non-Expressive & 20.35 & 2.86 &  2.267 &  0.0625 \\ 
     &      Yes & Expressive & 18.2   & 2.87 & 2.530 & 0.300 \\ 
    \midrule
     Seperation output  &  No & non-Expressive&  17.11  & 2.89 &  2.096 & 0.0172\\ 
     w/o BG     &       No & Expressive & 15.22 & 2.90 & 2.45 &0.2536 \\ 
           &       Yes & non-Expressive &18.26 & 2.97  & 2.028 & 0.057 \\ 
           &       Yes & Expressive& 16.10 & 2.97 & 2.507 &  0.3132 \\ 

    \midrule
     Seperation output  & No  & non-Expressive & 12.17 & 3.01 & 2.263 & 0.0794 \\
    w BG     &       No & Expressive&  12.33 & 3.00 & 2.40 & 0.2519\\ 
           &       Yes & non-Expressive & 15.70 & 3.09 & 2.257 & 0.079\\ 
           &       Yes & Expressive & 13.35 & 3.09 & 2.45 &  0.323 \\ 
    \bottomrule
    \end{tabular}
    \vskip -0.15in
    \label{table:translation-benchmark}
\end{table*}

\subsection{Benchmarking speech translation}
Here, we test our speech translation model in  an end-to-end manner. We compare four different model configurations: (1) Non-finetuned S2T with non-Expressive T2S (baseline from ~\cite{streamspeech}), (2) Non-finetuned S2T with Expressive T2S,   (3) Finetuned S2T with Expressive T2S,  and (4) Finetuned S2T with Expressive T2S. The chunk size of all models is set to  960~ms. 

As a sanity check, we first conducted an evaluation of the non-finetuned model on the original CoVoST2-CVSS Fr-En testset. The non-finetuned and non-expressive model achieved an ASR-BLEU score of  21.59, which is comparable to numbers reported in  \cite{streamspeech}, while our non-finetuned and expressive model achieved an ASR-BLEU score of 21.77. Then we tested the performance of the four models on the separated speech from the testing synthetic mixture (3000 mixture samples with background noise and 3000 samples without background noise). Notably, the separated speech is a subset of the entire CoVoST2-CVSS Fr-En test set. We evaluated all four models on three types of speech input: (1) raw single speaker speech before mixing (used as a reference ground truth), (2) separated speech without background noise in the mixture, and (3) separated speech with background noise in the mixture. 

As shown in Table~\ref{table:translation-benchmark}, across all three types of speech, our finetuned and non-expressive model had the best ASR-BLEU performance. Compared with the non-finetuned version, there is a 0.81 improvement on the raw single speaker speech, 1.15 improvement on separated speech without background noise and 3.52 improvement on the separated speech with background noise. It shows the effectiveness of our finetuning method for separation. Compared with our finetuned and non-expressive model, the ASR-BLEU score for the expressive model is lower. This is because our expressive encoder is frozen, and we found that the distortion and noise of input speech (even raw French speech in CoVoST2 has some noise and distortion) will also be preserved in the generated translated speech. This in turn  causes a drop in  performance  during ASR-BLEU calculations. In consideration of the streaming latency, all of the four models have similar performance of around 3 seconds. With separated speech, the AL slightly increase by 0.1-0.2s. In the evaluation of expressive preservation, our finetuned and expressive model has the best performance across all three types of input speech. Finally, we note that the performance with background noise is lower because the performance of the upstream source separation model in this case, as shown in Table~\ref{table:separation_benchmark}, is also lower. Increasing the model size for source separation can improve the translation performance in the presence of noise.

\begin{table*}[t!]
    \centering
    \caption{{Benchmarking results  for different language pairs on the separation output. (chunk size = 960ms)}}
    \vskip -0.15in
    \begin{tabular}{lcccccc}
    \toprule
    Input Language  &  Finetune  & T2S    & ASR-BLEU & AL (s) & Auto-PCP & VSim \\
      &  for Separation     &  &  &  & &\\
    \midrule
     French  &  No & non-Expressive&  17.11  & 2.89 &  2.096 & 0.0172\\ 
         &       No & Expressive & 15.22 & 2.90 & 2.45 &0.2536 \\ 
           &       Yes & non-Expressive &18.26 & 2.97  & 2.028 & 0.057 \\ 
           &       Yes & Expressive& 16.10 & 2.97 & 2.507 &  0.3132 \\ 
    \midrule
     Spanish  &  No & non-Expressive&  16.86  & 3.32 &  2.08 & 0.025 \\ 
          &       No & Expressive & 14.78 & 3.30 & 2.30 & 0.245 \\ 
           &       Yes & non-Expressive &19.30 & 3.27  & 2.11 & 0.026 \\ 
           &       Yes & Expressive& 16.76 & 3.27 & 2.30 &  0.25 \\ 

    \midrule
     German  & No  & non-Expressive & 12.33 & 3.27 & 2.06 & 0.060 \\
         &       No & Expressive&  12.61 & 3.28 & 2.25 & 0.270\\ 
           &       Yes & non-Expressive & 16.78 & 3.25 & 2.15 & 0.019\\ 
           &       Yes & Expressive & 14.52 & 3.25 & 2.26 &  0.274 \\ 
    \bottomrule
    \end{tabular}
    \vskip -0.15in
    \label{table:multi_lang}
\end{table*}

{Publicly available speech translation datasets, such as CVSS, CoVost2, and SpeechMatrix~\cite{duquenne2022speechmatrix}, include extensive data for upto 136 language pairs . Our translation pipeline can be easily extended to these language pairs. To demonstrate this, we trained our system on two additional pairs: Spanish-to-English and German-to-English. For Spanish-to-English, we used dataset pairs from CVSS and CoVost2, while for German-to-English, we utilized pairs from CVSS, CoVost2, and SpeechMatrix. The data was divided into training, validation, and test sets with no overlap. We created synthetic mixtures of Spanish and German speech in the same manner as for French (code to be made public). The separated Spanish and German speech was then fed into four different translation models, and the translation metrics were measured as shown in Table~\ref{table:multi_lang}. For Spanish-to-English translation, our fine-tuning method improved the ASR-BLEU score by 2.44 and 1.98 for non-expressive and expressive translation, respectively. The expressive T2S module increased VSim from 0.025 to 0.26 and Auto-PCP from 2.11 to 2.30 for the fine-tuned model. Similarly, for German-to-English translation, our fine-tuning method increased the ASR-BLEU score by 4.45 and 1.91 for non-expressive and expressive translation, respectively. The expressive T2S module increased VSim from 0.019 to 0.274 and Auto-PCP from 2.15 to 2.26 for the fine-tuned model.} 

\subsection{Benchmarking binaural rendering}

We evaluated the performance of our spatial preservation methods  on  6,000  test set samples (3,000 without background noise and 3,000  with background noise).
The accuracy of spatial cue preservation is assessed by measuring the differences in interaural time differences ($\Delta ITD$) and interaural level differences ($\Delta ILD$) between the rendered  translated speech and the ground-truth binaural speech before mixing~\cite{han2020real}.  $\Delta ITD$ is computed using cross-correlation, limited  to ±1 ms, following~\cite{may2010probabilistic}.

In Table~\ref{tab:binural_render_benchmark}, we compare three methods for binaural rendering: (1) Channel Duplication: simply duplicating the single-channel output from the translation model as binaural output, (2) Generic HRTF: using convolution with HRTF based on the estimated AoA to generate the binaural output, and (3) Generic HRTF \& ILD compensation:  using convolution  with HRTF based on the estimated AoA, along with the ILD compensation algorithm described in~\xref{sec:binaural_rendering}. The generic HRTF used for binaural rendering is not part of the training, validation, or testing sets. The Generic HRTF \& ILD Compensation method achieves the best performance with 72.3 µs $\Delta ITD$ and 0.162 dB $\Delta ILD$ without background noise, and 70.7 µs $\Delta ITD$ and 0.213 dB  $\Delta ILD$ with background noise. Note that using the generic HRTF alone achieves good results for ITD but performs poorly for ILD.

\begin{table}[b!]
    \centering
        \vskip -0.26in
    \caption{{Binaural rendering benchmark results.}}
    \vskip -0.15in
    \begin{tabular}{lcccc}
    \toprule
    Method & BG & $\Delta$ITD (µs) & $\Delta$ILD (dB)  \\
    \midrule
     Channel Duplication & No & 314.9 & 1.83 \\
       & Yes & 314.9 & 1.84  \\
     \midrule
     Generic HRTF & No & 72.3 &  1.47 \\
       & Yes &  70.7 & 1.46  \\
     \midrule
     Generic HRTF \& & No  & 72.3 &  0.16  \\
    ILD compensation & Yes & 70.7 & 0.21 \\
    \bottomrule
    \end{tabular}
    \label{tab:binural_render_benchmark}
\end{table}

\begin{figure*}[t!]
\centering
\includegraphics[width=1.7\columnwidth]{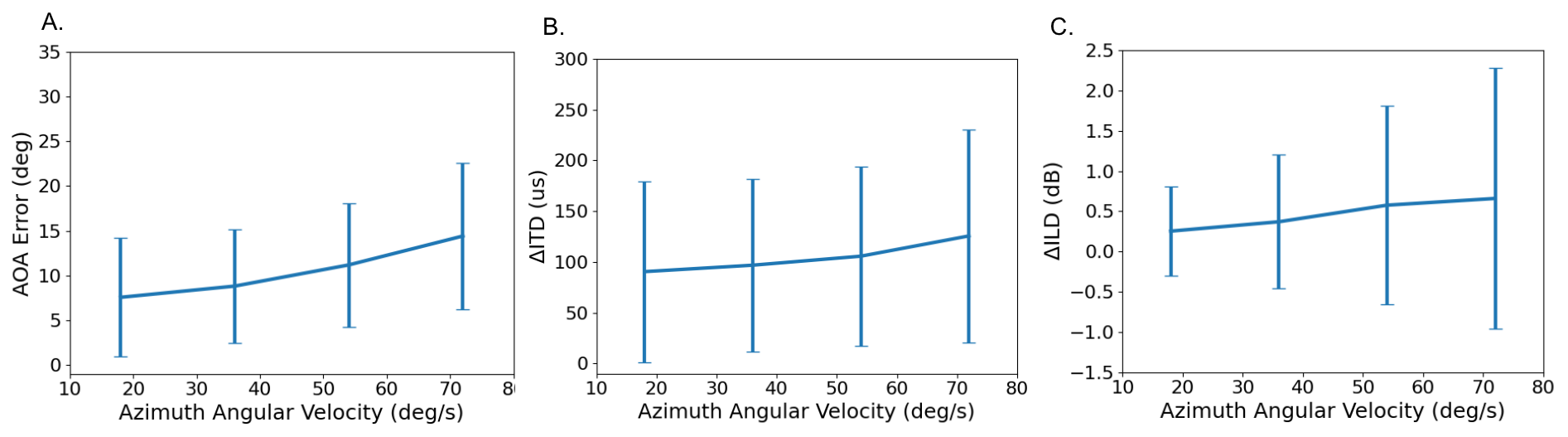}
\vskip -0.1in
\caption{{\bf {Binaural rendering  with motion.}} (A) AoA  error, (B) $\Delta ITD$, and (C) $\Delta ILD$  versus different azimuth angular speeds. }
\vskip -0.2in
\label{fig:motion}
\end{figure*}

Finally, we evaluate the performance of our spatial preservation method in the presence of motion. We simulate motion for a sound source using the BRIR dataset, following the steps in~\cite{tsh-chi24}. The motion experiments were conducted using only the CIPIC dataset, as it offers higher azimuth and elevation angular resolution compared to the other BRIR datasets, providing smoother motion simulation. In the simulations, a French speech source starts from an initial position with a randomly sampled azimuth and zero elevation. We then generate an array of positions over time with a time step of 50 ms and a constant azimuth angular velocity between $[-\pi/2, \pi/2]$ rad/s. Since the impulse responses in the CIPIC dataset are sampled at discrete spatial points, we use a nearest-neighbor approximation, selecting the BRIR from the dataset that is closest to the desired azimuth and polar angle at each time step in the motion trajectory. The Steam Audio SDK~\cite{steamaudio-sdk} is used to simulate the motion trajectory and obtain the binaural recording. Fig.~\ref{fig:motion}A shows that the AoA estimation is robust to different azimuth angular velocities. In Fig.~\ref{fig:motion}B and C, we compute the  $\Delta ILD$  and $\Delta ILD$  between each input binaural French speech chunk and the rendered English speech chunk at the same timestamp. While there is some degradation, our binaural rendering method still maintains good spatial preservation in the presence of motion. Fine-tuning the model with mobility data could further improve these results.

\section{Limitations and Discussion}\label{sec:limits}

{

In the real-world and synthetic metric evaluation, the Finetuned and Expressive model have lower ASR-BLEU than the Finetuned and Non-Expressive model. The primary reason is that the expressive encoder and vocoder are frozen in our training pipeline, leading to the preservation of noise and distortion in the output speech.
Finetuning these components requires large amounts of high-quality speech~\cite{seamlessexpressive}, which is not the main scope of our spatial translation. 
In the future, we can apply finetuning and knowledge distillation techniques to enable the expressive encoder and vocoder to robustly extract clean embeddings and produce clean speech in the presence of noise and distortion in the input audio. 

}

{The translation model used in this work has only around 175 million parameters, limiting its  BLEU score   and average lagging (AL). In contrast, models like SeamlessM4T, with around 2 billion parameters, offer significantly better translation capabilities. As mobile AI hardware accelerators advance, exploring larger models to enhance translation accuracy and reduce latency in spatial translation is a promising direction.  Regardless of the translation backbone, our spatial translation pipeline can be always applicable to support simultaneous multi-speaker translation and translation in noisy, real-world environments.}


{Some participants suggested that an ideal speech translation system should be customizable and adaptive to different use cases. In casual and social situations, such as travel, dating, and conversation, low-latency simultaneous translation is essential for smooth communication. However, for high-stakes applications like healthcare, law, immigration, and policing~\cite{medicine, law, huamnfactors-3}, translation accuracy takes precedence over translation latency. Hence, domain-specific calibration and personalization of the translation model are necessary. Since translations are not always perfect~\cite{humanfactors-1, humanfactors-2, huamnfactors-3}, techniques like back-translation and cross-validation can be used to improve user trust in AI-based speech translation.} 
  



Integrating our spatial translation system with binaural wireless earbuds would enable a wireless form factor. The spatial translation models do not need to run on the earbuds themselves but can operate on a nearby edge device, such as a smartphone or laptop, as Bluetooth round-trip latencies can be under 100 ms. However, this requires wirelessly streaming synchronized binaural audio from both earbuds, a capability demonstrated in prior work~\cite{clearbuds}. While our goal is spatial speech translation on hearables, as shown in~\xref{sec:translate:design}, our model can also output text in the target language. This could be used to display transcripts spatially on a visual display, aligned with the speakers' directions. Evaluating this on AR/VR headsets would be an interesting  research direction.
\section{Conclusion}

As we envision the future of real-time speech translation for hearables and augmented reality devices, incorporating spatial awareness becomes imperative. Our paper makes a important advance towards this goal by integrating spatial perception into real-time translation for hearables. We demonstrate binaural hearables that transform the wearer's auditory space into their native language while preserving spatial cues and unique voice characteristics in the binaural output. To achieve this goal, we make technical contributions across  source separation, machine translation and spatial computing. Our in-the-wild evaluations confirm generalization to real-world, unseen environments and participants.

\section*{Acknowledgments}

The researchers are partly supported
by the Moore Inventor Fellow award \#10617, Thomas J. Cable Endowed Professorship, and a UW CoMotion innovation gap fund. We  thank Jerimy Carroll for help with the conceptual figures in the paper.  We also thank Bohan Wu for his help with the initial dataset and model exploration This work was facilitated through the use of computational, storage, and networking infrastructure provided by the UW HYAK Consortium.

\bibliographystyle{ACM-Reference-Format}
\bibliography{paper}


\begin{thebibliography}{82}


\ifx \showCODEN    \undefined \def \showCODEN     #1{\unskip}     \fi
\ifx \showDOI      \undefined \def \showDOI       #1{#1}\fi
\ifx \showISBNx    \undefined \def \showISBNx     #1{\unskip}     \fi
\ifx \showISBNxiii \undefined \def \showISBNxiii  #1{\unskip}     \fi
\ifx \showISSN     \undefined \def \showISSN      #1{\unskip}     \fi
\ifx \showLCCN     \undefined \def \showLCCN      #1{\unskip}     \fi
\ifx \shownote     \undefined \def \shownote      #1{#1}          \fi
\ifx \showarticletitle \undefined \def \showarticletitle #1{#1}   \fi
\ifx \showURL      \undefined \def \showURL       {\relax}        \fi
\providecommand\bibfield[2]{#2}
\providecommand\bibinfo[2]{#2}
\providecommand\natexlab[1]{#1}
\providecommand\showeprint[2][]{arXiv:#2}

\bibitem[Agranovich et~al\mbox{.}(2024)]%
        {google-simul}
\bibfield{author}{\bibinfo{person}{Alex Agranovich}, \bibinfo{person}{Eliya Nachmani}, \bibinfo{person}{Oleg Rybakov}, \bibinfo{person}{Yifan Ding}, \bibinfo{person}{Ye Jia}, \bibinfo{person}{Nadav Bar}, \bibinfo{person}{Heiga Zen}, {and} \bibinfo{person}{Michelle~Tadmor Ramanovich}.} \bibinfo{year}{2024}\natexlab{}.
\newblock \bibinfo{title}{SimulTron: On-Device Simultaneous Speech to Speech Translation}.
\newblock
\newblock
\showeprint[arxiv]{2406.02133}~[eess.AS]
\urldef\tempurl%
\url{https://arxiv.org/abs/2406.02133}
\showURL{%
\tempurl}


\bibitem[Alexandrovsky et~al\mbox{.}(2020)]%
        {humanfactors-1}
\bibfield{author}{\bibinfo{person}{Dmitry Alexandrovsky}, \bibinfo{person}{Susanne Putze}, \bibinfo{person}{Michael Bonfert}, \bibinfo{person}{Sebastian H{\"o}ffner}, \bibinfo{person}{Pitt Michelmann}, \bibinfo{person}{Dirk Wenig}, \bibinfo{person}{Rainer Malaka}, {and} \bibinfo{person}{Jan~David Smeddinck}.} \bibinfo{year}{2020}\natexlab{}.
\newblock \showarticletitle{Unmet Needs and Opportunities for Mobile Translation AI}.
\newblock \bibinfo{journal}{\emph{CHI}} (\bibinfo{year}{2020}).
\newblock


\bibitem[Algazi et~al\mbox{.}(2001)]%
        {CIPIC}
\bibfield{author}{\bibinfo{person}{V.R. Algazi}, \bibinfo{person}{R.O. Duda}, \bibinfo{person}{D.M. Thompson}, {and} \bibinfo{person}{C. Avendano}.} \bibinfo{year}{2001}\natexlab{}.
\newblock \bibinfo{title}{The CIPIC HRTF database}.
\newblock , \bibinfo{numpages}{99-102}~pages.
\newblock
\urldef\tempurl%
\url{https://doi.org/10.1109/ASPAA.2001.969552}
\showDOI{\tempurl}


\bibitem[Apple(2024)]%
        {appleheadscan}
\bibfield{author}{\bibinfo{person}{Apple}.} \bibinfo{year}{2024}\natexlab{}.
\newblock \bibinfo{title}{Listen with Personalized Spatial Audio for AirPods and Beats}.
\newblock
\newblock
\urldef\tempurl%
\url{https://support.apple.com/en-us/102596}
\showURL{%
\tempurl}


\bibitem[Araki et~al\mbox{.}(2007)]%
        {bse}
\bibfield{author}{\bibinfo{person}{Shoko Araki}, \bibinfo{person}{Hiroshi Sawada}, {and} \bibinfo{person}{Shoji Makino}.} \bibinfo{year}{2007}\natexlab{}.
\newblock \showarticletitle{Blind Speech Separation in a Meeting Situation with Maximum SNR Beamformers}. In \bibinfo{booktitle}{\emph{ICASSP}}.
\newblock


\bibitem[Arivazhagan et~al\mbox{.}(2019)]%
        {learn2}
\bibfield{author}{\bibinfo{person}{Naveen Arivazhagan}, \bibinfo{person}{Colin Cherry}, \bibinfo{person}{Wolfgang Macherey}, \bibinfo{person}{Chung-Cheng Chiu}, \bibinfo{person}{Semih Yavuz}, \bibinfo{person}{Ruoming Pang}, \bibinfo{person}{Wei Li}, {and} \bibinfo{person}{Colin Raffel}.} \bibinfo{year}{2019}\natexlab{}.
\newblock \showarticletitle{Monotonic Infinite Lookback Attention for Simultaneous Machine Translation}. In \bibinfo{booktitle}{\emph{Proceedings of the 57th Annual Meeting of the Association for Computational Linguistics}}, \bibfield{editor}{\bibinfo{person}{Anna Korhonen}, \bibinfo{person}{David Traum}, {and} \bibinfo{person}{Llu{\'i}s M{\`a}rquez}} (Eds.).
\newblock


\bibitem[Babu et~al\mbox{.}(2022)]%
        {babu2021xls}
\bibfield{author}{\bibinfo{person}{Arun Babu}, \bibinfo{person}{Changhan Wang}, \bibinfo{person}{Andros Tjandra}, \bibinfo{person}{Kushal Lakhotia}, \bibinfo{person}{Qiantong Xu}, \bibinfo{person}{Naman Goyal}, \bibinfo{person}{Kritika Singh}, \bibinfo{person}{Patrick Von~Platen}, \bibinfo{person}{Yatharth Saraf}, \bibinfo{person}{Juan Pino}, {et~al\mbox{.}}} \bibinfo{year}{2022}\natexlab{}.
\newblock \showarticletitle{XLS-R: Self-supervised cross-lingual speech representation learning at scale}. In \bibinfo{booktitle}{\emph{InterSpeech}}.
\newblock


\bibitem[Bahdanau et~al\mbox{.}(2014)]%
        {Bahdanau2014NeuralMT}
\bibfield{author}{\bibinfo{person}{Dzmitry Bahdanau}, \bibinfo{person}{Kyunghyun Cho}, {and} \bibinfo{person}{Yoshua Bengio}.} \bibinfo{year}{2014}\natexlab{}.
\newblock \showarticletitle{Neural Machine Translation by Jointly Learning to Align and Translate}.
\newblock \bibinfo{journal}{\emph{CoRR}}  \bibinfo{volume}{abs/1409.0473} (\bibinfo{year}{2014}).
\newblock
\urldef\tempurl%
\url{https://api.semanticscholar.org/CorpusID:11212020}
\showURL{%
\tempurl}


\bibitem[Barrault et~al\mbox{.}(2025)]%
        {endtoend2}
\bibfield{author}{\bibinfo{person}{Loïc Barrault}, \bibinfo{person}{Yu-An Chung}, \bibinfo{person}{Mariano Meglioli}, \bibinfo{person}{David Dale}, \bibinfo{person}{Ning Dong}, \bibinfo{person}{Paul-Ambroise Duquenne}, \bibinfo{person}{Hady Elsahar}, \bibinfo{person}{Hongyu Gong}, \bibinfo{person}{Kevin Heffernan}, \bibinfo{person}{John Hoffman}, \bibinfo{person}{Christopher Klaiber}, \bibinfo{person}{Pengwei Li}, \bibinfo{person}{Daniel Licht}, \bibinfo{person}{Jean Maillard}, \bibinfo{person}{Alice Rakotoarison}, \bibinfo{person}{Kaushik Sadagopan}, \bibinfo{person}{Guillaume Wenzek}, \bibinfo{person}{Ethan Ye}, \bibinfo{person}{Bapi Akula}, {and} \bibinfo{person}{Skyler Wang}.} \bibinfo{year}{2025}\natexlab{}.
\newblock \showarticletitle{Joint speech and text machine translation for up to 100 languages}.
\newblock \bibinfo{journal}{\emph{Nature}}  \bibinfo{volume}{637} (\bibinfo{date}{01} \bibinfo{year}{2025}), \bibinfo{pages}{587--593}.
\newblock
\urldef\tempurl%
\url{https://doi.org/10.1038/s41586-024-08359-z}
\showDOI{\tempurl}


\bibitem[Blog(2024)]%
        {pixelbuds}
\bibfield{author}{\bibinfo{person}{Google Blog}.} \bibinfo{year}{2024}\natexlab{}.
\newblock \bibinfo{title}{Translate with Google Pixel Buds}.
\newblock
\newblock
\urldef\tempurl%
\url{https://support.google.com/googlepixelbuds/answer/7573100?hl=en}
\showURL{%
\tempurl}


\bibitem[Borsos et~al\mbox{.}(2023)]%
        {borsos2023audiolmlanguagemodelingapproach}
\bibfield{author}{\bibinfo{person}{Zalán Borsos}, \bibinfo{person}{Raphaël Marinier}, \bibinfo{person}{Damien Vincent}, \bibinfo{person}{Eugene Kharitonov}, \bibinfo{person}{Olivier Pietquin}, \bibinfo{person}{Matt Sharifi}, \bibinfo{person}{Dominik Roblek}, \bibinfo{person}{Olivier Teboul}, \bibinfo{person}{David Grangier}, \bibinfo{person}{Marco Tagliasacchi}, {and} \bibinfo{person}{Neil Zeghidour}.} \bibinfo{year}{2023}\natexlab{}.
\newblock \bibinfo{title}{AudioLM: a Language Modeling Approach to Audio Generation}.
\newblock
\newblock
\showeprint[arxiv]{2209.03143}~[cs.SD]


\bibitem[Carlini et~al\mbox{.}(2024)]%
        {carlini2024auditory}
\bibfield{author}{\bibinfo{person}{Alessandro Carlini}, \bibinfo{person}{Camille Bordeau}, {and} \bibinfo{person}{Maxime Ambard}.} \bibinfo{year}{2024}\natexlab{}.
\newblock \showarticletitle{Auditory localization: a comprehensive practical review}.
\newblock \bibinfo{journal}{\emph{Frontiers Psycholo.}} (\bibinfo{year}{2024}).
\newblock


\bibitem[Chatterjee et~al\mbox{.}(2022)]%
        {clearbuds}
\bibfield{author}{\bibinfo{person}{Ishan Chatterjee}, \bibinfo{person}{Maruchi Kim}, \bibinfo{person}{Vivek Jayaram}, \bibinfo{person}{Shyamnath Gollakota}, \bibinfo{person}{Ira Kemelmacher}, \bibinfo{person}{Shwetak Patel}, {and} \bibinfo{person}{Steven~M. Seitz}.} \bibinfo{year}{2022}\natexlab{}.
\newblock \showarticletitle{ClearBuds: wireless binaural earbuds for learning-based speech enhancement}. In \bibinfo{booktitle}{\emph{Proceedings of the 20th Annual International Conference on Mobile Systems, Applications and Services}} (Portland, Oregon) \emph{(\bibinfo{series}{MobiSys '22})}. \bibinfo{publisher}{Association for Computing Machinery}, \bibinfo{address}{New York, NY, USA}, \bibinfo{pages}{384–396}.
\newblock
\showISBNx{9781450391856}
\urldef\tempurl%
\url{https://doi.org/10.1145/3498361.3538933}
\showDOI{\tempurl}


\bibitem[Chen et~al\mbox{.}(2022)]%
        {Chen_2022}
\bibfield{author}{\bibinfo{person}{Sanyuan Chen}, \bibinfo{person}{Chengyi Wang}, \bibinfo{person}{Zhengyang Chen}, \bibinfo{person}{Yu Wu}, \bibinfo{person}{Shujie Liu}, \bibinfo{person}{Zhuo Chen}, \bibinfo{person}{Jinyu Li}, \bibinfo{person}{Naoyuki Kanda}, \bibinfo{person}{Takuya Yoshioka}, \bibinfo{person}{Xiong Xiao}, \bibinfo{person}{Jian Wu}, \bibinfo{person}{Long Zhou}, \bibinfo{person}{Shuo Ren}, \bibinfo{person}{Yanmin Qian}, \bibinfo{person}{Yao Qian}, \bibinfo{person}{Jian Wu}, \bibinfo{person}{Michael Zeng}, \bibinfo{person}{Xiangzhan Yu}, {and} \bibinfo{person}{Furu Wei}.} \bibinfo{year}{2022}\natexlab{}.
\newblock \showarticletitle{WavLM: Large-Scale Self-Supervised Pre-Training for Full Stack Speech Processing}.
\newblock \bibinfo{journal}{\emph{IEEE Journal of Selected Topics in Signal Processing}} \bibinfo{volume}{16}, \bibinfo{number}{6} (\bibinfo{date}{Oct.} \bibinfo{year}{2022}), \bibinfo{pages}{1505–1518}.
\newblock
\showISSN{1941-0484}
\urldef\tempurl%
\url{https://doi.org/10.1109/jstsp.2022.3188113}
\showDOI{\tempurl}


\bibitem[Chen et~al\mbox{.}(2024a)]%
        {soundbubble}
\bibfield{author}{\bibinfo{person}{Tuochao Chen}, \bibinfo{person}{Malek Itani}, \bibinfo{person}{Sefik Eskimez}, \bibinfo{person}{Takuya Yoshioka}, {and} \bibinfo{person}{Shyamnath Gollakota}.} \bibinfo{year}{2024}\natexlab{a}.
\newblock \showarticletitle{Hearable devices with sound bubbles}.
\newblock \bibinfo{journal}{\emph{Nature Electronics}}  \bibinfo{volume}{7} (\bibinfo{date}{11} \bibinfo{year}{2024}), \bibinfo{pages}{1047--1058}.
\newblock
\urldef\tempurl%
\url{https://doi.org/10.1038/s41928-024-01276-z}
\showDOI{\tempurl}


\bibitem[Chen et~al\mbox{.}(2024b)]%
        {tce}
\bibfield{author}{\bibinfo{person}{Tuochao Chen}, \bibinfo{person}{Qirui Wang}, \bibinfo{person}{Bohan Wu}, \bibinfo{person}{Malek Itani}, \bibinfo{person}{Sefik~Emre Eskimez}, \bibinfo{person}{Takuya Yoshioka}, {and} \bibinfo{person}{Shyamnath Gollakota}.} \bibinfo{year}{2024}\natexlab{b}.
\newblock \showarticletitle{Target conversation extraction: Source separation using turn-taking dynamics}. In \bibinfo{booktitle}{\emph{InterSpeech}}.
\newblock


\bibitem[Cheng et~al\mbox{.}(2024)]%
        {llmtrans}
\bibfield{author}{\bibinfo{person}{Shanbo Cheng}, \bibinfo{person}{Zhichao Huang}, \bibinfo{person}{Tom Ko}, \bibinfo{person}{Hang Li}, \bibinfo{person}{Ningxin Peng}, \bibinfo{person}{Lu Xu}, {and} \bibinfo{person}{Qini Zhang}.} \bibinfo{year}{2024}\natexlab{}.
\newblock \bibinfo{title}{Towards Achieving Human Parity on End-to-end Simultaneous Speech Translation via LLM Agent}.
\newblock
\newblock
\showeprint[arxiv]{2407.21646}~[cs.CL]
\urldef\tempurl%
\url{https://arxiv.org/abs/2407.21646}
\showURL{%
\tempurl}


\bibitem[Cho and Esipova(2016)]%
        {rule1}
\bibfield{author}{\bibinfo{person}{Kyunghyun Cho} {and} \bibinfo{person}{Masha Esipova}.} \bibinfo{year}{2016}\natexlab{}.
\newblock \bibinfo{title}{Can neural machine translation do simultaneous translation?}
\newblock
\newblock
\showeprint[arxiv]{1606.02012}~[cs.CL]


\bibitem[Communication et~al\mbox{.}(2023)]%
        {seamlessexpressive}
\bibfield{author}{\bibinfo{person}{Seamless Communication}, \bibinfo{person}{Loïc Barrault}, \bibinfo{person}{Yu-An Chung}, \bibinfo{person}{Mariano~Coria Meglioli}, \bibinfo{person}{David Dale}, \bibinfo{person}{Ning Dong}, \bibinfo{person}{Mark Duppenthaler}, \bibinfo{person}{Paul-Ambroise Duquenne}, \bibinfo{person}{Brian Ellis}, \bibinfo{person}{Hady Elsahar}, \bibinfo{person}{Justin Haaheim}, \bibinfo{person}{John Hoffman}, \bibinfo{person}{Min-Jae Hwang}, \bibinfo{person}{Hirofumi Inaguma}, \bibinfo{person}{Christopher Klaiber}, \bibinfo{person}{Ilia Kulikov}, \bibinfo{person}{Pengwei Li}, \bibinfo{person}{Daniel Licht}, \bibinfo{person}{Jean Maillard}, \bibinfo{person}{Ruslan Mavlyutov}, \bibinfo{person}{Alice Rakotoarison}, \bibinfo{person}{Kaushik~Ram Sadagopan}, \bibinfo{person}{Abinesh Ramakrishnan}, \bibinfo{person}{Tuan Tran}, \bibinfo{person}{Guillaume Wenzek}, \bibinfo{person}{Yilin Yang}, \bibinfo{person}{Ethan Ye}, \bibinfo{person}{Ivan Evtimov}, \bibinfo{person}{Pierre Fernandez},
  \bibinfo{person}{Cynthia Gao}, \bibinfo{person}{Prangthip Hansanti}, \bibinfo{person}{Elahe Kalbassi}, \bibinfo{person}{Amanda Kallet}, \bibinfo{person}{Artyom Kozhevnikov}, \bibinfo{person}{Gabriel~Mejia Gonzalez}, \bibinfo{person}{Robin~San Roman}, \bibinfo{person}{Christophe Touret}, \bibinfo{person}{Corinne Wong}, \bibinfo{person}{Carleigh Wood}, \bibinfo{person}{Bokai Yu}, \bibinfo{person}{Pierre Andrews}, \bibinfo{person}{Can Balioglu}, \bibinfo{person}{Peng-Jen Chen}, \bibinfo{person}{Marta~R. Costa-jussà}, \bibinfo{person}{Maha Elbayad}, \bibinfo{person}{Hongyu Gong}, \bibinfo{person}{Francisco Guzmán}, \bibinfo{person}{Kevin Heffernan}, \bibinfo{person}{Somya Jain}, \bibinfo{person}{Justine Kao}, \bibinfo{person}{Ann Lee}, \bibinfo{person}{Xutai Ma}, \bibinfo{person}{Alex Mourachko}, \bibinfo{person}{Benjamin Peloquin}, \bibinfo{person}{Juan Pino}, \bibinfo{person}{Sravya Popuri}, \bibinfo{person}{Christophe Ropers}, \bibinfo{person}{Safiyyah Saleem}, \bibinfo{person}{Holger Schwenk},
  \bibinfo{person}{Anna Sun}, \bibinfo{person}{Paden Tomasello}, \bibinfo{person}{Changhan Wang}, \bibinfo{person}{Jeff Wang}, \bibinfo{person}{Skyler Wang}, {and} \bibinfo{person}{Mary Williamson}.} \bibinfo{year}{2023}\natexlab{}.
\newblock \bibinfo{title}{Seamless: Multilingual Expressive and Streaming Speech Translation}.
\newblock
\newblock
\showeprint[arxiv]{2312.05187}~[cs.CL]
\urldef\tempurl%
\url{https://arxiv.org/abs/2312.05187}
\showURL{%
\tempurl}


\bibitem[Cornell et~al\mbox{.}(2023)]%
        {tse4}
\bibfield{author}{\bibinfo{person}{Samuele Cornell}, \bibinfo{person}{Zhong-Qiu Wang}, \bibinfo{person}{Yoshiki Masuyama}, \bibinfo{person}{Shinji Watanabe}, \bibinfo{person}{Manuel Pariente}, {and} \bibinfo{person}{Nobutaka Ono}.} \bibinfo{year}{2023}\natexlab{}.
\newblock \bibinfo{title}{Multi-Channel Target Speaker Extraction with Refinement: The WavLab Submission to the Second Clarity Enhancement Challenge}.
\newblock
\newblock
\showeprint[arxiv]{2302.07928}


\bibitem[Dalvi et~al\mbox{.}(2018)]%
        {rule2}
\bibfield{author}{\bibinfo{person}{Fahim Dalvi}, \bibinfo{person}{Nadir Durrani}, \bibinfo{person}{Hassan Sajjad}, {and} \bibinfo{person}{Stephan Vogel}.} \bibinfo{year}{2018}\natexlab{}.
\newblock \showarticletitle{Incremental Decoding and Training Methods for Simultaneous Translation in Neural Machine Translation}. In \bibinfo{booktitle}{\emph{Proceedings of the 2018 Conference of the North {A}merican Chapter of the Association for Computational Linguistics: Human Language Technologies, Volume 2 (Short Papers)}}, \bibfield{editor}{\bibinfo{person}{Marilyn Walker}, \bibinfo{person}{Heng Ji}, {and} \bibinfo{person}{Amanda Stent}} (Eds.).
\newblock


\bibitem[Dong et~al\mbox{.}(2023)]%
        {polyvoice}
\bibfield{author}{\bibinfo{person}{Qianqian Dong}, \bibinfo{person}{Zhiying Huang}, \bibinfo{person}{Qiao Tian}, \bibinfo{person}{Chen Xu}, \bibinfo{person}{Tom Ko}, \bibinfo{person}{Yunlong Zhao}, \bibinfo{person}{Siyuan Feng}, \bibinfo{person}{Tang Li}, \bibinfo{person}{Kexin Wang}, \bibinfo{person}{Xuxin Cheng}, \bibinfo{person}{Fengpeng Yue}, \bibinfo{person}{Ye Bai}, \bibinfo{person}{Xi Chen}, \bibinfo{person}{Lu Lu}, \bibinfo{person}{Zejun Ma}, \bibinfo{person}{Yuping Wang}, \bibinfo{person}{Mingxuan Wang}, {and} \bibinfo{person}{Yuxuan Wang}.} \bibinfo{year}{2023}\natexlab{}.
\newblock \bibinfo{title}{PolyVoice: Language Models for Speech to Speech Translation}.
\newblock
\newblock
\showeprint[arxiv]{2306.02982}~[cs.CL]
\urldef\tempurl%
\url{https://arxiv.org/abs/2306.02982}
\showURL{%
\tempurl}


\bibitem[Duquenne et~al\mbox{.}(2023)]%
        {duquenne2022speechmatrix}
\bibfield{author}{\bibinfo{person}{Paul-Ambroise Duquenne}, \bibinfo{person}{Hongyu Gong}, \bibinfo{person}{Ning Dong}, \bibinfo{person}{Jingfei Du}, \bibinfo{person}{Ann Lee}, \bibinfo{person}{Vedanuj Goswami}, \bibinfo{person}{Changhan Wang}, \bibinfo{person}{Juan Pino}, \bibinfo{person}{Beno{\^i}t Sagot}, {and} \bibinfo{person}{Holger Schwenk}.} \bibinfo{year}{2023}\natexlab{}.
\newblock \showarticletitle{{S}peech{M}atrix: A Large-Scale Mined Corpus of Multilingual Speech-to-Speech Translations}. In \bibinfo{booktitle}{\emph{Proceedings of the 61st Annual Meeting of the Association for Computational Linguistics}}.
\newblock


\bibitem[Elbayad et~al\mbox{.}(2020)]%
        {rule4}
\bibfield{author}{\bibinfo{person}{Maha Elbayad}, \bibinfo{person}{Laurent Besacier}, {and} \bibinfo{person}{Jakob Verbeek}.} \bibinfo{year}{2020}\natexlab{}.
\newblock \showarticletitle{Efficient Wait-k Models for Simultaneous Machine Translation}. In \bibinfo{booktitle}{\emph{InterSpeech}}.
\newblock


\bibitem[Eskimez et~al\mbox{.}(2022)]%
        {tse1}
\bibfield{author}{\bibinfo{person}{Sefik~Emre Eskimez}, \bibinfo{person}{Takuya Yoshioka}, \bibinfo{person}{Huaming Wang}, \bibinfo{person}{Xiaofei Wang}, \bibinfo{person}{Zhuo Chen}, {and} \bibinfo{person}{Xuedong Huang}.} \bibinfo{year}{2022}\natexlab{}.
\newblock \showarticletitle{Personalized speech enhancement: new models and Comprehensive evaluation}. In \bibinfo{booktitle}{\emph{ICASSP}}.
\newblock


\bibitem[Fang et~al\mbox{.}(2024)]%
        {ctc}
\bibfield{author}{\bibinfo{person}{Qingkai Fang}, \bibinfo{person}{Zhengrui Ma}, \bibinfo{person}{Yan Zhou}, \bibinfo{person}{Min Zhang}, {and} \bibinfo{person}{Yang Feng}.} \bibinfo{year}{2024}\natexlab{}.
\newblock \showarticletitle{{CTC}-based Non-autoregressive Textless Speech-to-Speech Translation}. In \bibinfo{booktitle}{\emph{Findings of the Association for Computational Linguistics: ACL 2024}}.
\newblock


\bibitem[Giri et~al\mbox{.}(2021)]%
        {tse2}
\bibfield{author}{\bibinfo{person}{Ritwik Giri}, \bibinfo{person}{Shrikant Venkataramani}, \bibinfo{person}{Jean-Marc Valin}, \bibinfo{person}{Umut Isik}, {and} \bibinfo{person}{Arvindh Krishnaswamy}.} \bibinfo{year}{2021}\natexlab{}.
\newblock \showarticletitle{Personalized PercepNet: Real-time, Low-complexity Target Voice Separation and Enhancement}. In \bibinfo{booktitle}{\emph{InterSpeech}}.
\newblock


\bibitem[Gu et~al\mbox{.}(2017)]%
        {t2umodel}
\bibfield{author}{\bibinfo{person}{Jiatao Gu}, \bibinfo{person}{James Bradbury}, \bibinfo{person}{Caiming Xiong}, \bibinfo{person}{Victor~OK Li}, {and} \bibinfo{person}{Richard Socher}.} \bibinfo{year}{2017}\natexlab{}.
\newblock \showarticletitle{Non-autoregressive neural machine translation}.
\newblock  (\bibinfo{year}{2017}).
\newblock


\bibitem[Gulati et~al\mbox{.}(2020)]%
        {conformer}
\bibfield{author}{\bibinfo{person}{Anmol Gulati}, \bibinfo{person}{James Qin}, \bibinfo{person}{Chung-Cheng Chiu}, \bibinfo{person}{Niki Parmar}, \bibinfo{person}{Yu Zhang}, \bibinfo{person}{Jiahui Yu}, \bibinfo{person}{Wei Han}, \bibinfo{person}{Shibo Wang}, \bibinfo{person}{Zhengdong Zhang}, \bibinfo{person}{Yonghui Wu}, {and} \bibinfo{person}{Ruoming Pang}.} \bibinfo{year}{2020}\natexlab{}.
\newblock \showarticletitle{Conformer: Convolution-augmented Transformer for Speech Recognition}. In \bibinfo{booktitle}{\emph{InterSpeech}}.
\newblock


\bibitem[Han et~al\mbox{.}(2020)]%
        {han2020real}
\bibfield{author}{\bibinfo{person}{Cong Han}, \bibinfo{person}{Yi Luo}, {and} \bibinfo{person}{Nima Mesgarani}.} \bibinfo{year}{2020}\natexlab{}.
\newblock \showarticletitle{Real-time binaural speech separation with preserved spatial cues}. In \bibinfo{booktitle}{\emph{ICASSP)}}. IEEE.
\newblock


\bibitem[IoSR-Surrey(2016)]%
        {rrbrir}
\bibfield{author}{\bibinfo{person}{IoSR-Surrey}.} \bibinfo{year}{2016}\natexlab{}.
\newblock \bibinfo{title}{IoSR-surrey/realroombrirs: Binaural impulse responses captured in real rooms.}
\newblock \bibinfo{howpublished}{\url{https://github.com/IoSR-Surrey/RealRoomBRIRs}}.
\newblock


\bibitem[IoSR-Surrey(2023)]%
        {CATT_RIR}
\bibfield{author}{\bibinfo{person}{IoSR-Surrey}.} \bibinfo{year}{2023}\natexlab{}.
\newblock \bibinfo{title}{Simulated Room Impulse Responses.}
\newblock \bibinfo{howpublished}{\url{https://iosr.uk/software/index.php}}.
\newblock


\bibitem[Itani et~al\mbox{.}(2023)]%
        {acousticswarm}
\bibfield{author}{\bibinfo{person}{Malek Itani}, \bibinfo{person}{Tuochao Chen}, \bibinfo{person}{Takuya Yoshioka}, {and} \bibinfo{person}{Shyamnath Gollakota}.} \bibinfo{year}{2023}\natexlab{}.
\newblock \showarticletitle{Creating speech zones with self-distributing acoustic swarms}.
\newblock \bibinfo{journal}{\emph{Nature Communications}}  \bibinfo{volume}{14} (\bibinfo{date}{09} \bibinfo{year}{2023}).
\newblock
\urldef\tempurl%
\url{https://doi.org/10.1038/s41467-023-40869-8}
\showDOI{\tempurl}


\bibitem[Jayaram et~al\mbox{.}(2023)]%
        {hrtfuist}
\bibfield{author}{\bibinfo{person}{Vivek Jayaram}, \bibinfo{person}{Ira Kemelmacher-Shlizerman}, {and} \bibinfo{person}{Steven~M. Seitz}.} \bibinfo{year}{2023}\natexlab{}.
\newblock \showarticletitle{HRTF Estimation in the Wild}. In \bibinfo{booktitle}{\emph{UIST}}. \bibinfo{publisher}{ACM}.
\newblock
\showISBNx{9798400701320}


\bibitem[Jenrungrot et~al\mbox{.}(2020)]%
        {cone-of-silence}
\bibfield{author}{\bibinfo{person}{Teerapat Jenrungrot}, \bibinfo{person}{Vivek Jayaram}, \bibinfo{person}{Steve Seitz}, {and} \bibinfo{person}{Ira Kemelmacher-Shlizerman}.} \bibinfo{year}{2020}\natexlab{}.
\newblock \showarticletitle{The Cone of Silence: Speech Separation by Localization}. In \bibinfo{booktitle}{\emph{Advances in Neural Information Processing Systems}}.
\newblock


\bibitem[Kit and Wong(2008)]%
        {law}
\bibfield{author}{\bibinfo{person}{Chunyu Kit} {and} \bibinfo{person}{Tak~Ming Wong}.} \bibinfo{year}{2008}\natexlab{}.
\newblock \showarticletitle{Comparative Evaluation of Online Machine Translation Systems with Legal Texts}.
\newblock \bibinfo{journal}{\emph{Law Library Journal}}  \bibinfo{volume}{100} (\bibinfo{year}{2008}), \bibinfo{pages}{299--321}.
\newblock
\urldef\tempurl%
\url{https://api.semanticscholar.org/CorpusID:13481458}
\showURL{%
\tempurl}


\bibitem[Kong et~al\mbox{.}(2020)]%
        {kong2020hifi}
\bibfield{author}{\bibinfo{person}{Jungil Kong}, \bibinfo{person}{Jaehyeon Kim}, {and} \bibinfo{person}{Jaekyoung Bae}.} \bibinfo{year}{2020}\natexlab{}.
\newblock \showarticletitle{Hifi-gan: Generative adversarial networks for efficient and high fidelity speech synthesis}.
\newblock \bibinfo{journal}{\emph{Advances in neural information processing systems}}  \bibinfo{volume}{33} (\bibinfo{year}{2020}), \bibinfo{pages}{17022--17033}.
\newblock


\bibitem[Le et~al\mbox{.}(2024)]%
        {voicebox}
\bibfield{author}{\bibinfo{person}{Matthew Le}, \bibinfo{person}{Apoorv Vyas}, \bibinfo{person}{Bowen Shi}, \bibinfo{person}{Brian Karrer}, \bibinfo{person}{Leda Sari}, \bibinfo{person}{Rashel Moritz}, \bibinfo{person}{Mary Williamson}, \bibinfo{person}{Vimal Manohar}, \bibinfo{person}{Yossi Adi}, \bibinfo{person}{Jay Mahadeokar}, {and} \bibinfo{person}{Wei-Ning Hsu}.} \bibinfo{year}{2024}\natexlab{}.
\newblock \showarticletitle{Voicebox: text-guided multilingual universal speech generation at scale}. In \bibinfo{booktitle}{\emph{Proceedings of the 37th International Conference on Neural Information Processing Systems}}.
\newblock


\bibitem[Lebert(2022)]%
        {classic}
\bibfield{author}{\bibinfo{person}{Marie Lebert}.} \bibinfo{year}{2022}\natexlab{}.
\newblock \bibinfo{title}{A short history of translation through the ages}.
\newblock
\newblock
\urldef\tempurl%
\url{https://www.iapti.org/iaptiarticle/a-short-history-of-translation-through-the-ages-marie-lebert-2/}
\showURL{%
\tempurl}


\bibitem[Lee et~al\mbox{.}(2022)]%
        {Lee2021DirectST}
\bibfield{author}{\bibinfo{person}{Ann Lee}, \bibinfo{person}{Peng-Jen Chen}, \bibinfo{person}{Changhan Wang}, \bibinfo{person}{Jiatao Gu}, \bibinfo{person}{Sravya Popuri}, \bibinfo{person}{Xutai Ma}, \bibinfo{person}{Adam Polyak}, \bibinfo{person}{Yossi Adi}, \bibinfo{person}{Qing He}, \bibinfo{person}{Yun Tang}, \bibinfo{person}{Juan Pino}, {and} \bibinfo{person}{Wei-Ning Hsu}.} \bibinfo{year}{2022}\natexlab{}.
\newblock \showarticletitle{Direct Speech-to-Speech Translation With Discrete Units}. In \bibinfo{booktitle}{\emph{Proceedings of the 60th Annual Meeting of the Association for Computational Linguistics}}, \bibfield{editor}{\bibinfo{person}{Smaranda Muresan}, \bibinfo{person}{Preslav Nakov}, {and} \bibinfo{person}{Aline Villavicencio}} (Eds.).
\newblock


\bibitem[Lei et~al\mbox{.}(2023)]%
        {denoise2}
\bibfield{author}{\bibinfo{person}{Tong Lei}, \bibinfo{person}{Zhongshu Hou}, \bibinfo{person}{Yuxiang Hu}, \bibinfo{person}{Wanyu Yang}, \bibinfo{person}{Tianchi Sun}, \bibinfo{person}{Xiaobin Rong}, \bibinfo{person}{Dahan Wang}, \bibinfo{person}{Kai Chen}, {and} \bibinfo{person}{Jing Lu}.} \bibinfo{year}{2023}\natexlab{}.
\newblock \showarticletitle{A Low-Latency Hybrid Multi-Channel Speech Enhancement System For Hearing Aids}. In \bibinfo{booktitle}{\emph{ICASSP}}. \bibinfo{pages}{1--2}.
\newblock


\bibitem[Levinson(2016)]%
        {Levinson2016TurntakingIH}
\bibfield{author}{\bibinfo{person}{Stephen~C. Levinson}.} \bibinfo{year}{2016}\natexlab{}.
\newblock \showarticletitle{Turn-taking in Human Communication – Origins and Implications for Language Processing}.
\newblock \bibinfo{journal}{\emph{Trends in Cog. Sci.}} (\bibinfo{year}{2016}).
\newblock


\bibitem[Liebling et~al\mbox{.}(2022)]%
        {humanfactors-2}
\bibfield{author}{\bibinfo{person}{Daniel~J. Liebling}, \bibinfo{person}{Katherine Heller}, \bibinfo{person}{Samantha Robertson}, {and} \bibinfo{person}{Wesley~Hanwen Deng}.} \bibinfo{year}{2022}\natexlab{}.
\newblock \showarticletitle{Opportunities for Human-centered Evaluation of Machine Translation Systems}. In \bibinfo{booktitle}{\emph{NAACL-HLT}}.
\newblock


\bibitem[Ma et~al\mbox{.}(2019)]%
        {rule3}
\bibfield{author}{\bibinfo{person}{Mingbo Ma}, \bibinfo{person}{Liang Huang}, \bibinfo{person}{Hao Xiong}, \bibinfo{person}{Renjie Zheng}, \bibinfo{person}{Kaibo Liu}, \bibinfo{person}{Baigong Zheng}, \bibinfo{person}{Chuanqiang Zhang}, \bibinfo{person}{Zhongjun He}, \bibinfo{person}{Hairong Liu}, \bibinfo{person}{Xing Li}, \bibinfo{person}{Hua Wu}, {and} \bibinfo{person}{Haifeng Wang}.} \bibinfo{year}{2019}\natexlab{}.
\newblock \showarticletitle{{STACL}: Simultaneous Translation with Implicit Anticipation and Controllable Latency using Prefix-to-Prefix Framework}. In \bibinfo{booktitle}{\emph{Proceedings of the 57th Annual Meeting of the Association for Computational Linguistics}}, \bibfield{editor}{\bibinfo{person}{Anna Korhonen}, \bibinfo{person}{David Traum}, {and} \bibinfo{person}{Llu{\'i}s M{\`a}rquez}} (Eds.). \bibinfo{publisher}{Association for Computational Linguistics}.
\newblock


\bibitem[Ma et~al\mbox{.}(2020)]%
        {learn1}
\bibfield{author}{\bibinfo{person}{Xutai Ma}, \bibinfo{person}{Juan Pino}, \bibinfo{person}{James Cross}, \bibinfo{person}{Liezl Puzon}, {and} \bibinfo{person}{Jiatao Gu}.} \bibinfo{year}{2020}\natexlab{}.
\newblock \showarticletitle{Monotonic Multihead Attention}. In \bibinfo{booktitle}{\emph{ICLR}}.
\newblock


\bibitem[Matusov et~al\mbox{.}(2005)]%
        {cascade2}
\bibfield{author}{\bibinfo{person}{Evgeny Matusov}, \bibinfo{person}{Stephan Kanthak}, {and} \bibinfo{person}{Hermann Ney}.} \bibinfo{year}{2005}\natexlab{}.
\newblock \showarticletitle{On the Integration of Speech Recognition and Statistical Machine Translation}. \bibinfo{pages}{3177--3180}.
\newblock
\urldef\tempurl%
\url{https://doi.org/10.21437/Interspeech.2005-726}
\showDOI{\tempurl}


\bibitem[May et~al\mbox{.}(2010)]%
        {may2010probabilistic}
\bibfield{author}{\bibinfo{person}{Tobias May}, \bibinfo{person}{Steven Van De~Par}, {and} \bibinfo{person}{Armin Kohlrausch}.} \bibinfo{year}{2010}\natexlab{}.
\newblock \showarticletitle{A probabilistic model for robust localization based on a binaural auditory front-end}.
\newblock \bibinfo{journal}{\emph{IEEE Transactions on audio, speech, and language processing}} \bibinfo{volume}{19}, \bibinfo{number}{1} (\bibinfo{year}{2010}), \bibinfo{pages}{1--13}.
\newblock


\bibitem[Mymanu(2024)]%
        {manu}
\bibfield{author}{\bibinfo{person}{Mymanu}.} \bibinfo{year}{2024}\natexlab{}.
\newblock \bibinfo{title}{Mymanu Click S}.
\newblock
\newblock
\urldef\tempurl%
\url{https://mymanu.com/products/mymanu-clik-s}
\showURL{%
\tempurl}


\bibitem[Ney(1999)]%
        {cascade1}
\bibfield{author}{\bibinfo{person}{H. Ney}.} \bibinfo{year}{1999}\natexlab{}.
\newblock \showarticletitle{Speech translation: coupling of recognition and translation}. In \bibinfo{booktitle}{\emph{ICASSP99}}, Vol.~\bibinfo{volume}{1}.
\newblock


\bibitem[{NLLB Team} et~al\mbox{.}(2022)]%
        {stopes}
\bibfield{author}{\bibinfo{person}{{NLLB Team}}, \bibinfo{person}{Marta~R. Costa-jussà}, \bibinfo{person}{James Cross}, \bibinfo{person}{Onur Çelebi}, \bibinfo{person}{Maha Elbayad}, \bibinfo{person}{Kenneth Heafield}, \bibinfo{person}{Kevin Heffernan}, \bibinfo{person}{Elahe Kalbassi}, \bibinfo{person}{Janice Lam}, \bibinfo{person}{Daniel Licht}, \bibinfo{person}{Jean Maillard}, \bibinfo{person}{Anna Sun}, \bibinfo{person}{Skyler Wang}, \bibinfo{person}{Guillaume Wenzek}, \bibinfo{person}{Al Youngblood}, \bibinfo{person}{Bapi Akula}, \bibinfo{person}{Loic Barrault}, \bibinfo{person}{Gabriel Mejia-Gonzalez}, \bibinfo{person}{Prangthip Hansanti}, \bibinfo{person}{John Hoffman}, \bibinfo{person}{Semarley Jarrett}, \bibinfo{person}{Kaushik~Ram Sadagopan}, \bibinfo{person}{Dirk Rowe}, \bibinfo{person}{Shannon Spruit}, \bibinfo{person}{Chau Tran}, \bibinfo{person}{Pierre Andrews}, \bibinfo{person}{Necip~Fazil Ayan}, \bibinfo{person}{Shruti Bhosale}, \bibinfo{person}{Sergey Edunov}, \bibinfo{person}{Angela Fan},
  \bibinfo{person}{Cynthia Gao}, \bibinfo{person}{Vedanuj Goswami}, \bibinfo{person}{Francisco Guzmán}, \bibinfo{person}{Philipp Koehn}, \bibinfo{person}{Alexandre Mourachko}, \bibinfo{person}{Christophe Ropers}, \bibinfo{person}{Safiyyah Saleem}, \bibinfo{person}{Holger Schwenk}, {and} \bibinfo{person}{Jeff Wang}.} \bibinfo{year}{2022}\natexlab{}.
\newblock \showarticletitle{No Language Left Behind: Scaling Human-Centered Machine Translation}.
\newblock  (\bibinfo{year}{2022}).
\newblock


\bibitem[NPR(2023)]%
        {startrek-npr}
\bibfield{author}{\bibinfo{person}{NPR}.} \bibinfo{year}{2023}\natexlab{}.
\newblock \bibinfo{title}{Finding your place in the galaxy with the help of Star Trek}.
\newblock
\newblock
\urldef\tempurl%
\url{https://www.npr.org/2023/10/14/1205714903/star-trek}
\showURL{%
\tempurl}


\bibitem[NPR(1998)]%
        {nprbabelfish}
\bibfield{author}{\bibinfo{person}{All Things~Considered NPR}.} \bibinfo{year}{1998}\natexlab{}.
\newblock \bibinfo{title}{Babelfish, a Translator Inspired by 'The Hitchhiker's Guide'}.
\newblock
\newblock
\urldef\tempurl%
\url{https://www.npr.org/1998/02/12/1036190/babelfish-a-translator-inspired-by-the-hitchhikers-guide}
\showURL{%
\tempurl}


\bibitem[Nunes~Vieira et~al\mbox{.}(2020)]%
        {medicine}
\bibfield{author}{\bibinfo{person}{Lucas Nunes~Vieira}, \bibinfo{person}{Minako O'Hagan}, {and} \bibinfo{person}{Carol O'Sullivan}.} \bibinfo{year}{2020}\natexlab{}.
\newblock \showarticletitle{Understanding the societal impacts of machine translation: a critical review of the literature on medical and legal use cases}.
\newblock \bibinfo{journal}{\emph{Information Communication and Society}} (\bibinfo{date}{06} \bibinfo{year}{2020}).
\newblock
\urldef\tempurl%
\url{https://doi.org/10.1080/1369118X.2020.1776370}
\showDOI{\tempurl}


\bibitem[Popuri et~al\mbox{.}(2022)]%
        {popuri2022enhanced}
\bibfield{author}{\bibinfo{person}{Sravya Popuri}, \bibinfo{person}{Peng-Jen Chen}, \bibinfo{person}{Changhan Wang}, \bibinfo{person}{Juan Pino}, \bibinfo{person}{Yossi Adi}, \bibinfo{person}{Jiatao Gu}, \bibinfo{person}{Wei-Ning Hsu}, {and} \bibinfo{person}{Ann Lee}.} \bibinfo{year}{2022}\natexlab{}.
\newblock \showarticletitle{Enhanced direct speech-to-speech translation using self-supervised pre-training and data augmentation}. In \bibinfo{booktitle}{\emph{InterSpeech}}.
\newblock


\bibitem[Post et~al\mbox{.}(2013)]%
        {cascade3}
\bibfield{author}{\bibinfo{person}{Matt Post}, \bibinfo{person}{G.~Santhosh Kumar}, \bibinfo{person}{Adam Lopez}, \bibinfo{person}{Damianos~G. Karakos}, \bibinfo{person}{Chris Callison-Burch}, {and} \bibinfo{person}{Sanjeev Khudanpur}.} \bibinfo{year}{2013}\natexlab{}.
\newblock \showarticletitle{Improved speech-to-text translation with the Fisher and Callhome Spanish-English speech translation corpus}. In \bibinfo{booktitle}{\emph{International Workshop on Spoken Language Translation}}.
\newblock


\bibitem[Raffel et~al\mbox{.}(2017)]%
        {learn3}
\bibfield{author}{\bibinfo{person}{Colin Raffel}, \bibinfo{person}{Minh-Thang Luong}, \bibinfo{person}{Peter~J. Liu}, \bibinfo{person}{Ron~J. Weiss}, {and} \bibinfo{person}{Douglas Eck}.} \bibinfo{year}{2017}\natexlab{}.
\newblock \showarticletitle{Online and linear-time attention by enforcing monotonic alignments}. In \bibinfo{booktitle}{\emph{Proceedings of the 34th International Conference on Machine Learning - Volume 70}} (Sydney, NSW, Australia) \emph{(\bibinfo{series}{ICML'17})}. \bibinfo{publisher}{JMLR.org}, \bibinfo{pages}{2837–2846}.
\newblock


\bibitem[Ren et~al\mbox{.}(2020)]%
        {simulspeech}
\bibfield{author}{\bibinfo{person}{Yi Ren}, \bibinfo{person}{Jinglin Liu}, \bibinfo{person}{Xu Tan}, \bibinfo{person}{Chen Zhang}, \bibinfo{person}{Tao Qin}, \bibinfo{person}{Zhou Zhao}, {and} \bibinfo{person}{Tie-Yan Liu}.} \bibinfo{year}{2020}\natexlab{}.
\newblock \showarticletitle{SimulSpeech: End-to-End Simultaneous Speech to Text Translation}. In \bibinfo{booktitle}{\emph{Annual Meeting of the Association for Computational Linguistics}}.
\newblock


\bibitem[Rethage et~al\mbox{.}(2018)]%
        {denoise1}
\bibfield{author}{\bibinfo{person}{Dario Rethage}, \bibinfo{person}{Jordi Pons}, {and} \bibinfo{person}{Xavier Serra}.} \bibinfo{year}{2018}\natexlab{}.
\newblock \showarticletitle{A Wavenet for Speech Denoising}. In \bibinfo{booktitle}{\emph{ICASSP}}.
\newblock


\bibitem[Robertson et~al\mbox{.}(2021)]%
        {huamnfactors-3}
\bibfield{editor}{\bibinfo{person}{Samantha Robertson}, \bibinfo{person}{Wesley Deng}, \bibinfo{person}{Timnit Gebru}, \bibinfo{person}{Margaret Mitchell}, \bibinfo{person}{Daniel~J. Liebling}, \bibinfo{person}{Michal Lahav}, \bibinfo{person}{Katherine Heller}, \bibinfo{person}{Mark Díaz}, \bibinfo{person}{Samy Bengio}, {and} \bibinfo{person}{Niloufar Salehi}} (Eds.). \bibinfo{year}{2021}\natexlab{}.
\newblock \bibinfo{booktitle}{\emph{Three Directions for the Design of Human-Centered Machine Translation}}.
\newblock


\bibitem[Rubenstein et~al\mbox{.}(2023)]%
        {audiopalm}
\bibfield{author}{\bibinfo{person}{Paul~K. Rubenstein}, \bibinfo{person}{Chulayuth Asawaroengchai}, \bibinfo{person}{Duc~Dung Nguyen}, \bibinfo{person}{Ankur Bapna}, \bibinfo{person}{Zalán Borsos}, \bibinfo{person}{Félix de Chaumont~Quitry}, \bibinfo{person}{Peter Chen}, \bibinfo{person}{Dalia~El Badawy}, \bibinfo{person}{Wei Han}, \bibinfo{person}{Eugene Kharitonov}, \bibinfo{person}{Hannah Muckenhirn}, \bibinfo{person}{Dirk Padfield}, \bibinfo{person}{James Qin}, \bibinfo{person}{Danny Rozenberg}, \bibinfo{person}{Tara Sainath}, \bibinfo{person}{Johan Schalkwyk}, \bibinfo{person}{Matt Sharifi}, \bibinfo{person}{Michelle~Tadmor Ramanovich}, \bibinfo{person}{Marco Tagliasacchi}, \bibinfo{person}{Alexandru Tudor}, \bibinfo{person}{M Velimirović}, \bibinfo{person}{Damien Vincent}, \bibinfo{person}{Jiahui Yu}, \bibinfo{person}{Y Wang}, \bibinfo{person}{Vicky Zayats}, \bibinfo{person}{N Zeghidour}, \bibinfo{person}{Yu Zhang}, \bibinfo{person}{Zhishuai Zhang}, \bibinfo{person}{Lukas Zilka}, {and}
  \bibinfo{person}{Christian Frank}.} \bibinfo{year}{2023}\natexlab{}.
\newblock \bibinfo{title}{AudioPaLM: A Large Language Model That Can Speak and Listen}.
\newblock
\newblock


\bibitem[SDK(2023)]%
        {steamaudio-sdk}
\bibfield{author}{\bibinfo{person}{SDK}.} \bibinfo{year}{2023}\natexlab{}.
\newblock \bibinfo{title}{Steam Audio}.
\newblock \bibinfo{howpublished}{\url{https://valvesoftware.github.io/steam-audio/}}.
\newblock


\bibitem[ShanonPearce(2022)]%
        {ShanonPearce}
\bibfield{author}{\bibinfo{person}{ShanonPearce}.} \bibinfo{year}{2022}\natexlab{}.
\newblock \bibinfo{title}{Shanonpearce/ash-listening-set: A dataset of filters for headphone correction and binaural synthesis of spatial audio systems on headphones}.
\newblock
\newblock
\urldef\tempurl%
\url{https://github.com/ShanonPearce/ASH-Listening-Set/tree/main}
\showURL{%
\tempurl}


\bibitem[Shen et~al\mbox{.}(2023)]%
        {shen2023naturalspeech2latentdiffusion}
\bibfield{author}{\bibinfo{person}{Kai Shen}, \bibinfo{person}{Zeqian Ju}, \bibinfo{person}{Xu Tan}, \bibinfo{person}{Yanqing Liu}, \bibinfo{person}{Yichong Leng}, \bibinfo{person}{Lei He}, \bibinfo{person}{Tao Qin}, \bibinfo{person}{Sheng Zhao}, {and} \bibinfo{person}{Jiang Bian}.} \bibinfo{year}{2023}\natexlab{}.
\newblock \bibinfo{title}{NaturalSpeech 2: Latent Diffusion Models are Natural and Zero-Shot Speech and Singing Synthesizers}.
\newblock
\newblock
\showeprint[arxiv]{2304.09116}~[eess.AS]
\urldef\tempurl%
\url{https://arxiv.org/abs/2304.09116}
\showURL{%
\tempurl}


\bibitem[Sperber and Paulik(2020)]%
        {endtoend1}
\bibfield{author}{\bibinfo{person}{Matthias Sperber} {and} \bibinfo{person}{Matthias Paulik}.} \bibinfo{year}{2020}\natexlab{}.
\newblock \showarticletitle{Speech Translation and the End-to-End Promise: Taking Stock of Where We Are}. In \bibinfo{booktitle}{\emph{Annual Meeting of the Association for Computational Linguistics}}.
\newblock


\bibitem[Subakan et~al\mbox{.}(2021)]%
        {better1}
\bibfield{author}{\bibinfo{person}{Cem Subakan}, \bibinfo{person}{Mirco Ravanelli}, \bibinfo{person}{Samuele Cornell}, \bibinfo{person}{Mirko Bronzi}, {and} \bibinfo{person}{Jianyuan Zhong}.} \bibinfo{year}{2021}\natexlab{}.
\newblock \showarticletitle{Attention Is All You Need In Speech Separation}. In \bibinfo{booktitle}{\emph{ICASSP 2021}}.
\newblock


\bibitem[Timekettle({[n.\,d.]})]%
        {timekettle}
\bibfield{author}{\bibinfo{person}{Timekettle}.} \bibinfo{year}{[n.\,d.]}\natexlab{}.
\newblock \bibinfo{title}{Timekettle WT2 Edge/W3 Real-time Translator Earbuds, 2-way simultaneous interpretation}.
\newblock
\newblock
\urldef\tempurl%
\url{https://www.timekettle.co/products/wt2-edge-online-voice-language-translator-earbuds}
\showURL{%
\tempurl}


\bibitem[Vaswani et~al\mbox{.}(2017)]%
        {attentionpaper}
\bibfield{author}{\bibinfo{person}{Ashish Vaswani}, \bibinfo{person}{Noam Shazeer}, \bibinfo{person}{Niki Parmar}, \bibinfo{person}{Jakob Uszkoreit}, \bibinfo{person}{Llion Jones}, \bibinfo{person}{Aidan~N Gomez}, \bibinfo{person}{\L~ukasz Kaiser}, {and} \bibinfo{person}{Illia Polosukhin}.} \bibinfo{year}{2017}\natexlab{}.
\newblock \showarticletitle{Attention is All you Need}. In \bibinfo{booktitle}{\emph{Advances in Neural Information Processing Systems}}, Vol.~\bibinfo{volume}{30}.
\newblock


\bibitem[Veluri et~al\mbox{.}(2023a)]%
        {better2}
\bibfield{author}{\bibinfo{person}{Bandhav Veluri}, \bibinfo{person}{Justin Chan}, \bibinfo{person}{Malek Itani}, \bibinfo{person}{Tuochao Chen}, \bibinfo{person}{Takuya Yoshioka}, {and} \bibinfo{person}{Shyamnath Gollakota}.} \bibinfo{year}{2023}\natexlab{a}.
\newblock \showarticletitle{Real-Time Target Sound Extraction}. In \bibinfo{booktitle}{\emph{ICASSP}}. \bibinfo{pages}{1--5}.
\newblock


\bibitem[Veluri et~al\mbox{.}(2023b)]%
        {semantichearing}
\bibfield{author}{\bibinfo{person}{Bandhav Veluri}, \bibinfo{person}{Malek Itani}, \bibinfo{person}{Justin Chan}, \bibinfo{person}{Takuya Yoshioka}, {and} \bibinfo{person}{Shyamnath Gollakota}.} \bibinfo{year}{2023}\natexlab{b}.
\newblock \showarticletitle{Semantic Hearing: Programming Acoustic Scenes with Binaural Hearables}. In \bibinfo{booktitle}{\emph{Proceedings of the 36th Annual ACM Symposium on User Interface Software and Technology}} (San Francisco, CA, USA) \emph{(\bibinfo{series}{UIST '23})}. \bibinfo{publisher}{Association for Computing Machinery}, \bibinfo{address}{New York, NY, USA}, Article \bibinfo{articleno}{89}, \bibinfo{numpages}{15}~pages.
\newblock
\showISBNx{9798400701320}
\urldef\tempurl%
\url{https://doi.org/10.1145/3586183.3606779}
\showDOI{\tempurl}


\bibitem[Veluri et~al\mbox{.}(2024)]%
        {tsh-chi24}
\bibfield{author}{\bibinfo{person}{Bandhav Veluri}, \bibinfo{person}{Malek Itani}, \bibinfo{person}{Tuochao Chen}, \bibinfo{person}{Takuya Yoshioka}, {and} \bibinfo{person}{Shyamnath Gollakota}.} \bibinfo{year}{2024}\natexlab{}.
\newblock \showarticletitle{Look Once to Hear: Target Speech Hearing with Noisy Examples}. In \bibinfo{booktitle}{\emph{CHI}} (Honolulu, HI, USA) \emph{(\bibinfo{series}{CHI '24})}. \bibinfo{publisher}{ACM}, Article \bibinfo{articleno}{37}, \bibinfo{numpages}{16}~pages.
\newblock
\showISBNx{9798400703300}


\bibitem[Wang et~al\mbox{.}(2022a)]%
        {directionalhearing1}
\bibfield{author}{\bibinfo{person}{Anran Wang}, \bibinfo{person}{Maruchi Kim}, \bibinfo{person}{Hao Zhang}, {and} \bibinfo{person}{Shyamnath Gollakota}.} \bibinfo{year}{2022}\natexlab{a}.
\newblock \showarticletitle{Hybrid Neural Networks for On-device Directional Hearing}.
\newblock \bibinfo{journal}{\emph{AAAI}} (\bibinfo{year}{2022}).
\newblock


\bibitem[Wang et~al\mbox{.}(2023)]%
        {wang2023neuralcodeclanguagemodels}
\bibfield{author}{\bibinfo{person}{Chengyi Wang}, \bibinfo{person}{Sanyuan Chen}, \bibinfo{person}{Yu Wu}, \bibinfo{person}{Ziqiang Zhang}, \bibinfo{person}{Long Zhou}, \bibinfo{person}{Shujie Liu}, \bibinfo{person}{Zhuo Chen}, \bibinfo{person}{Yanqing Liu}, \bibinfo{person}{Huaming Wang}, \bibinfo{person}{Jinyu Li}, \bibinfo{person}{Lei He}, \bibinfo{person}{Sheng Zhao}, {and} \bibinfo{person}{Furu Wei}.} \bibinfo{year}{2023}\natexlab{}.
\newblock \bibinfo{title}{Neural Codec Language Models are Zero-Shot Text to Speech Synthesizers}.
\newblock
\newblock
\showeprint[arxiv]{2301.02111}~[cs.CL]


\bibitem[Wang et~al\mbox{.}(2022b)]%
        {endtoend4}
\bibfield{author}{\bibinfo{person}{Peidong Wang}, \bibinfo{person}{Eric Sun}, \bibinfo{person}{Jian Xue}, \bibinfo{person}{Yu Wu}, \bibinfo{person}{Long Zhou}, \bibinfo{person}{Yashesh Gaur}, \bibinfo{person}{Shujie Liu}, {and} \bibinfo{person}{Jinyu Li}.} \bibinfo{year}{2022}\natexlab{b}.
\newblock \showarticletitle{LAMASSU: A Streaming Language-Agnostic Multilingual Speech Recognition and Translation Model Using Neural Transducers}. In \bibinfo{booktitle}{\emph{Interspeech}}.
\newblock
\urldef\tempurl%
\url{https://api.semanticscholar.org/CorpusID:258968116}
\showURL{%
\tempurl}


\bibitem[Waverly(2024)]%
        {waverly}
\bibfield{author}{\bibinfo{person}{Waverly}.} \bibinfo{year}{2024}\natexlab{}.
\newblock \bibinfo{title}{Waverly labs Earbuds}.
\newblock
\newblock
\urldef\tempurl%
\url{https://www.waverlylabs.com/}
\showURL{%
\tempurl}


\bibitem[Wu et~al\mbox{.}(2016)]%
        {machinetranslate}
\bibfield{author}{\bibinfo{person}{Yonghui Wu}, \bibinfo{person}{Mike Schuster}, \bibinfo{person}{Zhifeng Chen}, \bibinfo{person}{Quoc~V. Le}, \bibinfo{person}{Mohammad Norouzi}, \bibinfo{person}{Wolfgang Macherey}, \bibinfo{person}{Maxim Krikun}, \bibinfo{person}{Yuan Cao}, \bibinfo{person}{Qin Gao}, \bibinfo{person}{Klaus Macherey}, \bibinfo{person}{Jeff Klingner}, \bibinfo{person}{Apurva Shah}, \bibinfo{person}{Melvin Johnson}, \bibinfo{person}{Xiaobing Liu}, \bibinfo{person}{Łukasz Kaiser}, \bibinfo{person}{Stephan Gouws}, \bibinfo{person}{Yoshikiyo Kato}, \bibinfo{person}{Taku Kudo}, \bibinfo{person}{Hideto Kazawa}, \bibinfo{person}{Keith Stevens}, \bibinfo{person}{George Kurian}, \bibinfo{person}{Nishant Patil}, \bibinfo{person}{Wei Wang}, \bibinfo{person}{Cliff Young}, \bibinfo{person}{Jason Smith}, \bibinfo{person}{Jason Riesa}, \bibinfo{person}{Alex Rudnick}, \bibinfo{person}{Oriol Vinyals}, \bibinfo{person}{Greg Corrado}, \bibinfo{person}{Macduff Hughes}, {and} \bibinfo{person}{Jeffrey Dean}.}
  \bibinfo{year}{2016}\natexlab{}.
\newblock \showarticletitle{Google's Neural Machine Translation System: Bridging the Gap between Human and Machine Translation}.
\newblock \bibinfo{journal}{\emph{CoRR}}  \bibinfo{volume}{abs/1609.08144} (\bibinfo{year}{2016}).
\newblock


\bibitem[Xu and Choudhury(2022)]%
        {romit1}
\bibfield{author}{\bibinfo{person}{Zhongweiyang Xu} {and} \bibinfo{person}{Romit~Roy Choudhury}.} \bibinfo{year}{2022}\natexlab{}.
\newblock \showarticletitle{Learning to Separate Voices by Spatial Regions}.
\newblock \bibinfo{journal}{\emph{ICML}} (\bibinfo{year}{2022}).
\newblock


\bibitem[Xue et~al\mbox{.}(2022)]%
        {endtoend3}
\bibfield{author}{\bibinfo{person}{Jian Xue}, \bibinfo{person}{Peidong Wang}, \bibinfo{person}{Jinyu Li}, \bibinfo{person}{Matt Post}, {and} \bibinfo{person}{Yashesh Gaur}.} \bibinfo{year}{2022}\natexlab{}.
\newblock \showarticletitle{Large-Scale Streaming End-to-End Speech Translation with Neural Transducers}. In \bibinfo{booktitle}{\emph{Interspeech}}.
\newblock


\bibitem[Xutai~Ma(2020)]%
        {simuleval2020}
\bibfield{author}{\bibinfo{person}{Changhan Wang Jiatao Gu Juan~Pino Xutai~Ma, Mohammad Javad~Dousti}.} \bibinfo{year}{2020}\natexlab{}.
\newblock \showarticletitle{Simuleval: An evaluation toolkit for simultaneous translation}. In \bibinfo{booktitle}{\emph{Proceedings of the EMNLP}}.
\newblock


\bibitem[Yang et~al\mbox{.}(2024)]%
        {takuya1}
\bibfield{author}{\bibinfo{person}{Mu Yang}, \bibinfo{person}{Naoyuki Kanda}, \bibinfo{person}{Xiaofei Wang}, \bibinfo{person}{Junkun Chen}, \bibinfo{person}{Peidong Wang}, \bibinfo{person}{Jian Xue}, \bibinfo{person}{Jinyu Li}, {and} \bibinfo{person}{Takuya Yoshioka}.} \bibinfo{year}{2024}\natexlab{}.
\newblock \showarticletitle{Diarist: Streaming Speech Translation with Speaker Diarization}. In \bibinfo{booktitle}{\emph{ICASSP}}. \bibinfo{pages}{10866--10870}.
\newblock


\bibitem[Zhang et~al\mbox{.}(2024)]%
        {streamspeech}
\bibfield{author}{\bibinfo{person}{Shaolei Zhang}, \bibinfo{person}{Qingkai Fang}, \bibinfo{person}{Shoutao Guo}, \bibinfo{person}{Zhengrui Ma}, \bibinfo{person}{Min Zhang}, {and} \bibinfo{person}{Yang Feng}.} \bibinfo{year}{2024}\natexlab{}.
\newblock \showarticletitle{{S}tream{S}peech: Simultaneous Speech-to-Speech Translation with Multi-task Learning}. In \bibinfo{booktitle}{\emph{Proceedings of the 62nd Annual Meeting of the Association for Computational Linguistics (Volume 1: Long Papers)}}.
\newblock


\bibitem[Zhang and Feng(2021)]%
        {rule5}
\bibfield{author}{\bibinfo{person}{Shaolei Zhang} {and} \bibinfo{person}{Yang Feng}.} \bibinfo{year}{2021}\natexlab{}.
\newblock \showarticletitle{Universal Simultaneous Machine Translation with Mixture-of-Experts Wait-k Policy}. In \bibinfo{booktitle}{\emph{Proceedings of the 2021 Conference on Empirical Methods in Natural Language Processing}}. \bibinfo{publisher}{Association for Computational Linguistics}.
\newblock


\bibitem[Žmolíková et~al\mbox{.}(2017)]%
        {tse3}
\bibfield{author}{\bibinfo{person}{Kateřina Žmolíková}, \bibinfo{person}{Marc Delcroix}, \bibinfo{person}{Keisuke Kinoshita}, \bibinfo{person}{Takuya Higuchi}, \bibinfo{person}{Atsunori Ogawa}, {and} \bibinfo{person}{Tomohiro Nakatani}.} \bibinfo{year}{2017}\natexlab{}.
\newblock \showarticletitle{{Speaker-Aware Neural Network Based Beamformer for Speaker Extraction in Speech Mixtures}}. In \bibinfo{booktitle}{\emph{Proc. Interspeech 2017}}.
\newblock


\end{thebibliography}

\end{document}